\documentclass[review,3p]{elsarticle}

\usepackage{floatrow}
\newfloatcommand{capbtabbox}{table}[][\FBwidth]

\usepackage[linesnumbered,ruled]{algorithm2e}

\usepackage{scalerel,stackengine}
\stackMath
\newcommand\predhat[1]{%
\savestack{\tmpbox}{\stretchto{%
  \scaleto{%
    \scalerel*[\widthof{\ensuremath{#1}}]{\kern-.6pt\bigwedge\kern-.6pt}%
    {\rule[-\textheight/2]{1ex}{\textheight}}
  }{\textheight}%
}{0.5ex}}%
\stackon[1pt]{#1}{\tmpbox}%
}

\usepackage{placeins}
\usepackage[table,x11names]{xcolor}
\definecolor{light-gray}{gray}{0.95}

\usepackage{caption, subcaption}
\usepackage{bm}
\usepackage{makecell}

\usepackage{amsfonts}
\usepackage{graphicx}
\usepackage{hyperref}
\usepackage{booktabs,tabularx}
\usepackage{float}
\usepackage{amsmath}
\usepackage{amstext}
\usepackage{color,soul}
\usepackage{multirow}
\usepackage{lscape}
\usepackage[title]{appendix}

\newcommand\numberthis{\addtocounter{equation}{1}\tag{\theequation}}

\newcommand{\fgrref}[1]{\figurename\ \ref{#1}}
\newcommand{\tblref}[1]{Table\ \ref{#1}}

\newcolumntype{P}[1]{>{\centering\arraybackslash}p{#1}}
\newcolumntype{Q}[1]{>{\raggedright\arraybackslash}p{#1}}
\newcolumntype{Z}{>{\raggedright\let\newline\\\arraybackslash\hspace{0pt}}X}

\newcolumntype{x}{>{\centering\arraybackslash}X}
\newcolumntype{M}[1]{>{\centering\arraybackslash}m{#1}}

\newcommand{\mtl}{MTL}
\newcommand{\icp}{ICP_{5-95}}
\newcommand{\mil}{MIL_{5-95}}

\newcommand{\aoD}{o \in \mathcal{N}}
\newcommand{\adD}{d \in \mathcal{N}}
\newcommand{\aodD}{o,d \in \mathcal{N}}
\newcommand{\ac}{1 \leq c \leq C}
\newcommand{\ak}{1 \leq k \leq K}
\newcommand{\q}{\quad}
\newcommand{\qq}{\qquad}

\newcommand{\inv}{^{\raisebox{.2ex}{$\scriptscriptstyle-\!1$}}}

\newenvironment{conditions*}
  {\par\vspace{\abovedisplayskip}\noindent
   \tabularx{\textwidth}{>{$}l<{$} @{${}:{}$} >{\raggedright\arraybackslash}X}}
  {\endtabularx\par\vspace{\belowdisplayskip}}
  
\usepackage{epstopdf}
\AppendGraphicsExtensions{.tif}

\DeclareMathOperator*{\mode}{mode}

\usepackage{tikz}
\usetikzlibrary{shapes,arrows}
\usetikzlibrary{positioning}
\usetikzlibrary{arrows.meta}
\usetikzlibrary{calc}

\graphicspath{{images/}}
\usepackage{lineno,hyperref}
\modulolinenumbers[5]

\usepackage{rotating, lipsum}
\definecolor{Gray}{gray}{0.85}
\newcolumntype{a}{>{\columncolor{Gray}}c}

\journal{Transportation Research Part B}

\begin{document}

\tikzset{%
  neuron/.style={
    circle,
    draw,
    minimum size=1cm
  },
  missing/.style={
    draw=none, 
    scale=2.5,
    text height=0.333cm,
    execute at begin node=\color{black}$\vdots$
  },
  missingsm/.style={
    draw=none, 
    scale=1,
    text height=0.333cm,
    execute at begin node=\color{black}$\vdots$
  },
  sqneuron/.style={
    draw,
    minimum size=1cm
  },
  cdots/.style={
    draw=none, 
    scale=3,
    text height=0.333cm,
    execute at begin node=\color{black}$\cdots$
  },  
}

\begin{frontmatter}

\title{Online Predictive Optimization Framework for Stochastic Demand-Responsive Transit Services}

\address[dtuaddress]{Technical University of Denmark (DTU), DTU Management, 2800 Kgs. Lyngby, Denmark}

\address[ntuaddress2]{Nanyang Technological University (NTU), Graduate College, 50 Nanyang Avenue, Singapore 637553}

\address[ntuaddress]{Nanyang Technological University (NTU), School of Electrical and Electronic Engineering, Singapore 639798}

\author[dtuaddress]{Inon Peled\corref{mycorrespondingauthor}}
\cortext[mycorrespondingauthor]{Corresponding author}
\ead{inonpe@dtu.dk}

\author[ntuaddress2]{Kelvin Lee}
\ead{kelvin003@e.ntu.edu.sg}

\author[dtuaddress]{Yu Jiang}
\ead{yujiang@dtu.dk}

\author[ntuaddress]{Justin Dauwels}
\ead{jdauwels@ntu.edu.sg}

\author[dtuaddress]{Francisco C. Pereira}
\ead{camara@dtu.dk}

\begin{abstract}
This study develops an online predictive optimization framework for dynamically operating a transit service in an area of crowd movements.
The proposed framework integrates demand prediction and supply optimization to periodically redesign the service routes based on recently observed demand.
To predict demand for the service, we use Quantile Regression to estimate the marginal distribution of movement counts between each pair of serviced locations. 
The framework then combines these marginals into a joint demand distribution by constructing a Gaussian copula, which captures the structure of correlation between the marginals.
For supply optimization, we devise a linear programming model, which simultaneously determines the route structure and the service frequency according to the predicted demand.
Importantly, our framework both preserves the uncertainty structure of future demand and leverages this for robust route optimization, while keeping both components decoupled.
We evaluate our framework using a real-world case study of autonomous mobility in a university campus in Denmark.
The results show that our framework often obtains the ground truth optimal solution, and can outperform conventional methods for route optimization, which do not leverage full predictive distributions.
\end{abstract}


\end{frontmatter}


\section{Introduction} \label{sec:introduction}

Various institutions around the world are increasingly incorporating autonomous vehicle fleets into their on-campus mobility solutions \cite{ntu2018autonomous, linc2018largest, driverless2018}.
Whereas traditional bus services often follow fixed itineraries, autonomous vehicle fleets can have their itineraries dynamically adapted in real-time, e.g., per changing demand for mobility during a work day.
In other words, autonomous mobility services are amenable to  demand-responsive routing, if demand can be predicted ahead of time.
Fortunately, institutional campuses are often finely meshed with WiFi access points, which can in turn be used as sensors of crowd presence \cite{sevtsuk2009mapping, meneses2012large}.
By using WiFi information to detect movements of wireless devices across campus, the current and future demand for mobility can be estimated.
An opportunity thus emerges for real-time, demand-responsive autonomous mobility. 

Existing literature in transport is rich with methods for demand prediction and route optimization.
However, as we present in Section \ref{sec:literature}, these studies often fall short of fully treating both these aspects. 
First, existing studies often concentrate either only on demand prediction or supply optimization, leaving the other to external works. 
Second, these studies often insufficiently account for uncertainty in demand and supply, and use point estimates instead of more informative full distributions.
In this study, however, we offer a framework that explicitly incorporates both demand prediction and supply optimization, while leveraging demand uncertainty to optimize demand-responsive transit services.

\subsection{Overview of Our Solution Framework} \label{sec:overview}

\begin{figure}[ht]
\centering
\includegraphics[width=\linewidth]{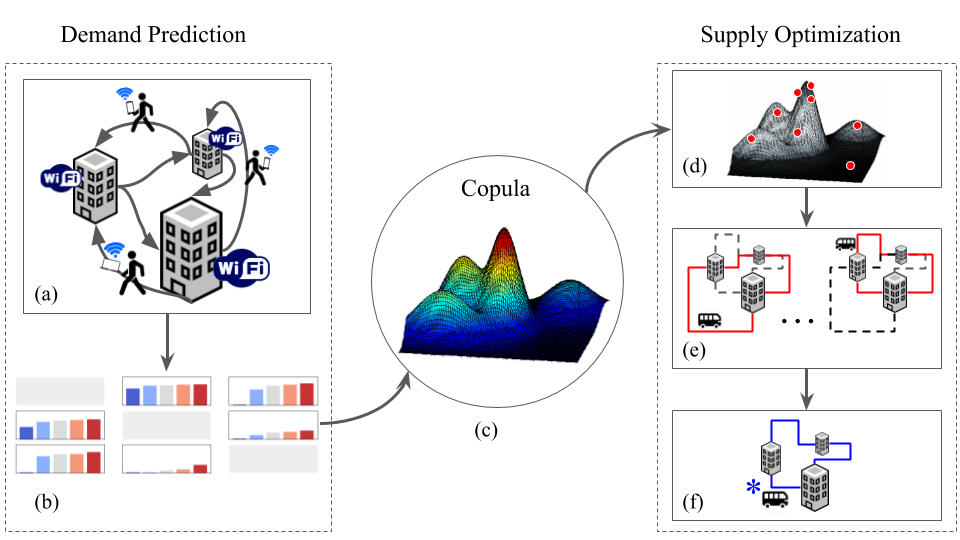}
\caption{Our online predictive optimization framework. (a) Data about crowd movements is collected via WiFi probing. (b) The marginal distribution of each OD pair is estimated through Quantile Regression. (c) A copula combines the marginal distributions into a joint demand distribution. (d) Samples are drawn from the joint distribution. (e) An optimal route and corresponding frequency is determined for each sample. (f) The most frequently obtained optimal solution is selected for operation.}
\label{fig:framework}
\end{figure}

The setting in which we apply our framework is that of a mobility service, for instance, a fleet of autonomous shuttles in a university or hospital.
People move in the serviceable area between connected locations, such as buildings connected by roads.
These movements are captured by a network of sensors, e.g., a network of WiFi hotspots probing people's wireless devices.
The different locations form a partition of the spatial dimension, and we further partition the temporal dimension into consecutive lags (e.g., $1 h$).

Crowd movements between different locations at different times are subject to \emph{uncertainty}.
For example, we may incompletely know the number and location of all people in the serviced area at each moment, and it is also uncertain when and why each of these people chooses to change location.
Equivalently, in probability theory terms, the number of people who move between locations over time follows some latent (i.e., unknown) spatio-temporal distribution -- a view also expressed in, e.g., \cite{gonzalez2008understanding}, \cite{guo2012discovering}, and \cite{guo2017novel}.
At each lag then, a latent \emph{joint} distribution accounts for the simultaneous movements between all OD pairs, each of which has its own latent \emph{marginal} distribution of movements from Origin to Destination.
The marginal distributions of different OD pairs may be correlated, and the joint distribution accounts for any such correlations.

The number of movements between each OD pair represents the \emph{demand} for the mobility service, while the fleet of vehicles represents the \emph{supply}.
Based on the data collected at previous lags, our online framework predicts the demand, and accordingly optimizes the supply for the next lag, as illustrated in \fgrref{fig:framework}.
For demand prediction, the framework estimates the marginal distributions through Quantile Regression.
For supply optimization, the framework decides an optimal design for the transit network using a scenario-based method.
In each scenario, a draw is sampled from the predictive joint distribution of demand, and a corresponding linear programming instance is solved to obtain route structure and frequency for the sampled demand.
Finally, the solutions for all samples are combined into an overall optimal network design for the next time lag.

This joint demand distribution is obtained by constructing a Gaussian copula, which is thus a key link between between demand prediction and supply optimization in our framework.
The use of a copula allows us to decouple these two components.
Furthermore, as the copula preserves all correlations between OD pairs in the joint distribution, we are free to work with marginal demand distributions, which are easier to fit accurately for each OD pair.

Importantly, our framework retains the uncertainty in travel demand (rather than reduce it to point estimates), and takes advantage of this uncertainty during supply optimization.
In this study, we advocate for preserving this uncertainty by estimating each marginal distribution through quantiles of its Cumulative Distribution Function.
Another advantage of our framework is that our scenario-based approach to supply optimization yields a robust optimization scheme. That is, whereas most existing scenario-based optimization methods use predefined sets of scenarios, we generate scenarios dynamically and demand-responsively, based on real-time demand prediction.

\subsection{Summary of Contributions}

The main contributions of this paper are thus as follows:
\begin{itemize}
\item We develop a online predictive optimization framework for the demand-responsive transit network design problem, which decouples demand prediction from adaptive supply optimization, and integrates them through a copula.

\item On the demand side, we develop several Quantile Regression models for predicting the marginal demand distribution for each Origin-Destination pair.

\item On the supply side, we devise a robust optimization method for demand-responsive transit network design under stochastic demand, based on a novel minimum cost flow formulation.

\item Using a case study based on real-world WiFi data, we demonstrate the capabilities of our predictive optimization framework, and show that it can outperform conventional optimization methods.
\end{itemize}

\subsection{Paper Structure}

The rest of this paper is structured as follows. 
In Section \ref{sec:literature}, we review literature on demand prediction and supply optimization under uncertainty, and indicate the advantages of our study over previous works. 
In Section \ref{s:dempred}, we present different models for demand prediction with uncertainty, which we offer to do through Quantile Regression.
There, we also apply the prediction models to a case study of autonomous mobility in a Danish university campus, based on actual WiFi records.
In Section \ref{sec:supplyopt}, we present a novel online method to determine the route structure and frequency for demand-responsive transit services under stochastic demand. 
In Section \ref{sec:case_sup}, we apply this stochastic optimization method to the aforementioned case study.
Finally, in Section \ref{sec:conclusion}, we summarize our findings, draw conclusions, and identify future research directions. 

\section{Literature Review} \label{sec:literature}

In this Section, we review existing studies and state-of-the-art solution methods, while comparing them to our solution methodology.
We begin by reviewing demand prediction with uncertainty, proceed to review supply optimization under demand uncertainty, and finally review methods for integrating both aspects.

\subsection{Demand Prediction with Uncertainty}

Accurate modeling of travel demand is essential for properly planning transit services: the more accurate the demand forecast, the better can service resources be allocated and scheduled ahead of time \cite{rodrigues2018beyond}.
Prediction of future demand for transport has thus been a long studied research topic, resulting in a plethora of parametric and non-parametric techniques for demand modeling  \cite{de2007uncertainty, rasouli2012uncertainty}.
However, despite the importance of accurate demand prediction, previous studies have often tended to over-simplify by providing only point estimates of future values \cite{rodrigues2018heteroscedastic}.
For example, transport studies often provide only the mean and standard deviation of the predictive distribution, either directly or, respectively, through the center and bounds of a confidence interval.
Such point estimate methods, e.g., Maximum Likelihood Estimation and Maximum A-Posteriori, are commonly applied to a wide variety of demand modeling techniques \cite{tsai2009neural, yang2013sensitivity, shao2014estimation, xue2015short}.

As the distribution of future demand may be highly irregular (e.g., skewed or multi-modal), reducing it to summary statistics may result in losing important information about the uncertainty structure of future demand. 
In turn, this can lead to inaccurate resource allocation and passenger dissatisfaction.
Indeed, there are several benefits for preserving uncertainty in predictions, rather than providing only point estimates \cite{rodrigues2018heteroscedastic, yang2019estimating, li2019robust}.
On one hand, preserving uncertainty conveys a high degree of confidence in the predictions, so that corresponding decisions can be made more intelligently.
For example, when given a full predictive distribution, a supply optimization method can prepare for a full range of possible scenarios, from best case to worst case. 
On the other hand, providing only point estimates might be misleading, e.g., the mean of a multi-modal distribution might in fact lie in a neighborhood of low probability.

To preserve uncertainty, Quantile Regression (QR) can be used to approximate a full predictive distribution by estimating several of its quantiles, without assuming any particular parametric form.
This method has been applied to various problems in, e.g., econometric analysis \cite{chen2019quantile}, weather forecasting \cite{khan2019prediction}, and transport modeling \cite{rodrigues2018beyond}.
The QR model itself can follow either of several functional forms, such as linear or splines \cite{koenker2005quantile}, non-linear or non-parametric with Gaussian Processes \cite{antunes2017review, yang2018power}, or vector-valued \cite{sangnier2016joint}. 
As more quantiles are used, the approximation which QR yields becomes more precise and more robust to artifacts in the true predictive distribution, such as multi-modality and non-symmetry.

In this paper, we evaluate the proposed framework through a case study, in which we estimate the distribution of future demand through Quantile Regression.
Nevertheless, the framework supports multiple models other than QR for predicting arbitrary marginal distributions.
For a moderately sized dataset as in our case study, Bayesian Inference \cite{peled2019model} can also be used with proper modeling of random variables and their dependency structure. 
Alternatively, Deep Neural Network models can be constructed along with multiple prediction intervals, as described in \cite{mazloumi2011prediction} and \cite{khosravi2011comprehensive}.

\subsection{Transit Network Design and Frequency Setting Problem (TNDFS)}

The transit route planning process typically consists of five steps, as outlined in \cite{Ceder1986-lk}. 
The process is sequential, in that the decisions at each step become the input for the next step. 
The process starts with a network design problem, where the bus stops and routes are decided.
Then the frequencies are determined based on the fleet size, followed by timetabling, vehicle scheduling, and finally the crew scheduling.

The transit network design and frequency setting problem (TNDFS) is a combination of two distinct sequential problems, i.e. the transit route network design problem (TRNDP), which deals with the planning of optimal routes for transit services, and the frequency determination problem (FDP), which determines the frequencies for each of those routes.
Extensive research into TRNDP variants and solution methods has been conducted since the late 1960s.
We refer the readers to \cite{Kepaptsoglou2009-sw} for an extensive review of previous studies on TRNDP up to year 2007, in which the author classified the studies according to their objective functions, decision variables, transit network structure, demand patterns and characteristics, as well as the solution methods. We also provide references \cite{ceder_2016, GUIHAIRE20081251, FARAHANI2013281} that address various other aspects of the problem.
Here, we focus the mathematical programming formulation for TRNDP under stochasticity. 

\subsubsection{Mathematical programming formulation for TRNDP} \label{sec:formulation-literature}
It is clear that while the solutions obtained at each step of the transit planning process are optimal, they may not necessarily be optimal for the overall transit planning problem.
Despite the temptation to formulate a single model to globally optimize the transit planning procedure, the complexity involved should be taken into account, as the problem at each step of the process is highly combinatorial.
The two steps relevant to our study are network design, and frequency determination, as we review next.

For network design, TRNDP instances are usually formulated based on a network graph, where the nodes represent transit stops, and edges represent connective paths between nodes.
The objective is then to select which transit stops to serve, and the order of visiting them in each of the routes, based on travel demands and generalized costs.
A solution method for network design gives the optimal routes and their corresponding temporal route length.

The frequency determination problem, on the other hand, is to optimally allocate vehicles to the different routes.
This allocation largely depends on the temporal route lengths. 
The integration of the two problems can increase the complexity substantially, and often results either in a single highly nonlinear model \cite{Cipriani2012-zm, Fan2008-lg, Fan_Wei2006-mj}, or a bi-level mixed integer model \cite{Szeto2011-qj, Szeto2014-pn}.
The resulting formulations are complex, and rely on heuristics to solve for suboptimal solutions.
In contrast, our optimization method uses a linear, multi-stage formulation, which can be solved through sampling, so that the solution is optimal in expectation.

\subsubsection{Stochastic Transit Network Design Problem with Uncertain Demands} \label{sec:stocTRNDP}

In the literature, TRNDP is usually formulated as a \emph{static} linear program with parameters that take on deterministic values.
In practice, however, these parameters are often not static but \emph{stochastic}, so that they follow some probability distributions.

While some previous studies have considered demand elasticity, which is an inherent property of a real transit network, most studies have only considered fixed demand.
In \cite{Fan_Wei2006-mj}, Fan and Machemehl have attributed this to the NP-hard complexity of TRNDP.
The consideration of elastic demands often results in an iterative procedure, which repeatedly chooses routes structure and demand splits until some convergence criterion is achieved \cite{Cipriani2012-yr, Lee_Young-Jae2005-on, Fan_Wei2006-lc}.
However, this would mean that no optimal solution can be guaranteed due to the heuristic nature of the problem.
Therefore, in our framework, demands are not considered as directly dependent on the service quality.
Instead, the elasticity is internalized in the demand stochasticity alongside other factors, e.g., weather and seasonal variations.

Approaches for dealing with such stochasticity commonly reduce the distributions to point estimates.
One such approach is to use the expected values of parameter distributions \cite{CIPRIANI20123, Fan2008-lg}.
While the expected value formulation may be simple to obtain, its solution nonetheless lacks robustness.
To overcome this weakness, minimax robust optimization can be used \cite{An2015-tv, Laporte2010-ud, Lou2009-et, An2016-bx}, whereby the maximum of the support of the parameters is taken instead of their expected values.
Doing so ensures that the solution is feasible for all possible combinations of parameters.
This approach is often called the \emph{worst-case scenario} approach, because the solution space encapsulates all possible combinations, including the worst-case scenario.

Nevertheless, there are drawbacks to both approaches for dealing with stochasticity in parameters.
First, by reducing the parameter distributions to point values, both approaches discard of useful information in the full distributions.
Second, while minimax robust optimization guarantees feasibility for all possible demand scenarios, it can sometimes be overly conservative, as acknowledged in \cite{An2015-tv}.
In contrast, we optimize service routes and frequencies based on a full estimate of the predictive demand distribution, rather than point estimates.

Another commonly used method to improve robustness is scenario-based robust optimization.
There, stochasticity is generated either by adding random perturbations to the average demands, by explicitly constructing demand scenarios for different seasons and/or time of the day, or by random sampling of parameters from their probability densities.
For instance, in \cite{Amiripour_S_M_Mahdi2014-om} Amiripour et al. consider TRNDP with variable demands.
To simulate demand stochasticity, they generate $480$ perturbations of a demand matrix, each by adding stochastic noise. 
While this method does add robustness to the solution by introducing random noise in the demand, the random perturbations may not necessarily reflect the stochasticity that can be observed in collected demand data.

In contrast, the optimization method we devise in this study takes advantage of the full predictive distributions of parameters. 
As (stochastic) TRNDP is an NP-hard combinatorial problem, solution methods for TRNDP are often approximate, and so is our method.
Specifically, our solution method for stochastic TRNDP is based on sampling from the distribution of parameters, so that each sample is a static problem instance, uniquely defined by the drawn parameter values.
Consequently, the solution that our method yields is optimal in expectation.

\subsection{Combining Demand Prediction with Supply Optimization}

Several studies cater for both aspects of demand-responsive supply optimization, i.e., explicitly provide both a model of future demand and an algorithm for demand-responsive supply optimization.
The methods that these studies develop for combining both aspects generally fall into two categories: hybrid and decoupled. 
Hybrid methods (e.g., \cite{Stein1978DAR, cortes2009hybrid, powell2003stochastic}) combine demand prediction and supply optimization within one formulation, often through stochastic programming.
Conversely, decoupled methods (e.g., \cite{ferrucci2013pro, alonso2017predictive, iglesias2018data}) model each of the two components separately, so that supply is optimized based on the output of an independent demand model.

As noted in \cite{ichoua2006exploiting}, notable works on hybrid methods include a line of papers by Powell and his team, dating back as early as 1988 (\cite{powell1988comparative, powell1996stochastic, godfrey2002adaptive, powell2003stochastic, spivey2004dynamic}). 
A more recent example of the hybrid approach is by Cortes et al. \cite{cortes2009hybrid} (2009), who study a delivery-and-pickup service with predetermined stations, and develop a state-space model where the objective function to be optimized consists of both demand estimate and route cost. 
Our solution framework, however, promotes a decoupled approach, which allows us to model each component independently, while joining them through a copula (as defined later in Sec. \ref{sec:copula}). 
This decoupling allows us to freely compare several models with uncertainty estimates on the demand prediction side, while offering a non-myopic algorithm on the supply optimization side, i.e., an algorithm that considers multiple future scenarios to improve user gains.
The use of a copula further allows us to derive marginal demand distributions for each OD pair, rather than estimate an entire joint distribution at once.

A common deficiency in works about decoupled methods is that the unified solution they propose relies on an possibly overly simplified demand prediction component.
Such is the case in the seminal work by Stein \cite{Stein1978DAR} (1978), who studies route optimization for Dial-a-Ride services, while assuming that users' requests are independently drawn from a uniform spatial distribution over the serviced area. 
Ferrucci et al. \cite{ferrucci2013pro} (2013) present another approach to demand-responsive supply optimization, whereby the service area is segmented, and each segment is assumed to follow a Poisson distribution with time-dependent rate.
Iglesias et al. \cite{iglesias2018data} (2018) offer a data-driven framework to rebalance an autonomous on-demand fleet, where demand is predicted based on a Deep Neural Network model, which does not yield uncertainty estimates.
While our framework also takes a decoupled approach, it does not presume a particular form of demand distribution, and can rather approximate it through e.g., Quantile Regression (QR, as defined in Sec. \ref{sec:qr}).
This is particularly applicable to the case study of mobility in a university campus, where student movements are often spatio-temporally correlated and unevenly distributed.

Let us conclude the literature review with two recent (2017) investigations that bear particular resemblance to ours. 
The first is by Alonso-Mora et al. \cite{alonso2017predictive}, where the authors study urban-scale route optimization for a hypothetical fleet of self-driving taxis. 
Similarly to our approach, they too leverage historical data for real-time demand prediction, and construct a marginal probability distribution for each OD pair.
Nevertheless, their method relies on the frequentist approach, which can be beneficial for large datasets, but fails to retain enough uncertainty for smaller datasets as in our case.


The second study which particularly resembles ours is by Miller et al. \cite{miller2017predictive}, where the authors too predictively optimize a small fleet of on-campus autonomous shuttles, with the objective of minimizing expected customer waiting time. 
However, whereas they focus on predicting key locations for proactive positioning of the vehicles, our objective is to predictively optimize service routes and frequencies.
Also, their sources of demand information are users' requests via a dedicated smartphone app, and sensors that the few vehicle carry around. 
In contrast, our source of demand information is a network of hundreds of WiFi hotspots, fixed all over campus, which thus provide high spatio-temporal observability.
Finally, whereas our framework offers online predictive optimization, their method is offline, and they differ online demand prediction to future work.

\section{Demand Prediction through Quantile Regression} \label{s:dempred}

In this Section, we describe how Quantile Regression (QR) can be applied to demand prediction with uncertainty. 
We demonstrate the application of QR step-by-step through a case study of predictive optimization for an autonomous shuttle service in a Danish University campus as part of a real-world project ("LINC", ref. \cite{linc2018largest}).
For this project, we obtained a dataset of WiFi records, collected from various buildings on the campus for several weeks in 2017 and 2018.
We also use this case study in later Sections to describe and test our stochastic optimization method.

\subsection{Data for Demand Estimation}

\begin{figure}[ht]
    \centering
    \begin{subfigure}[b]{0.28\textwidth}
        \includegraphics[width=\textwidth]{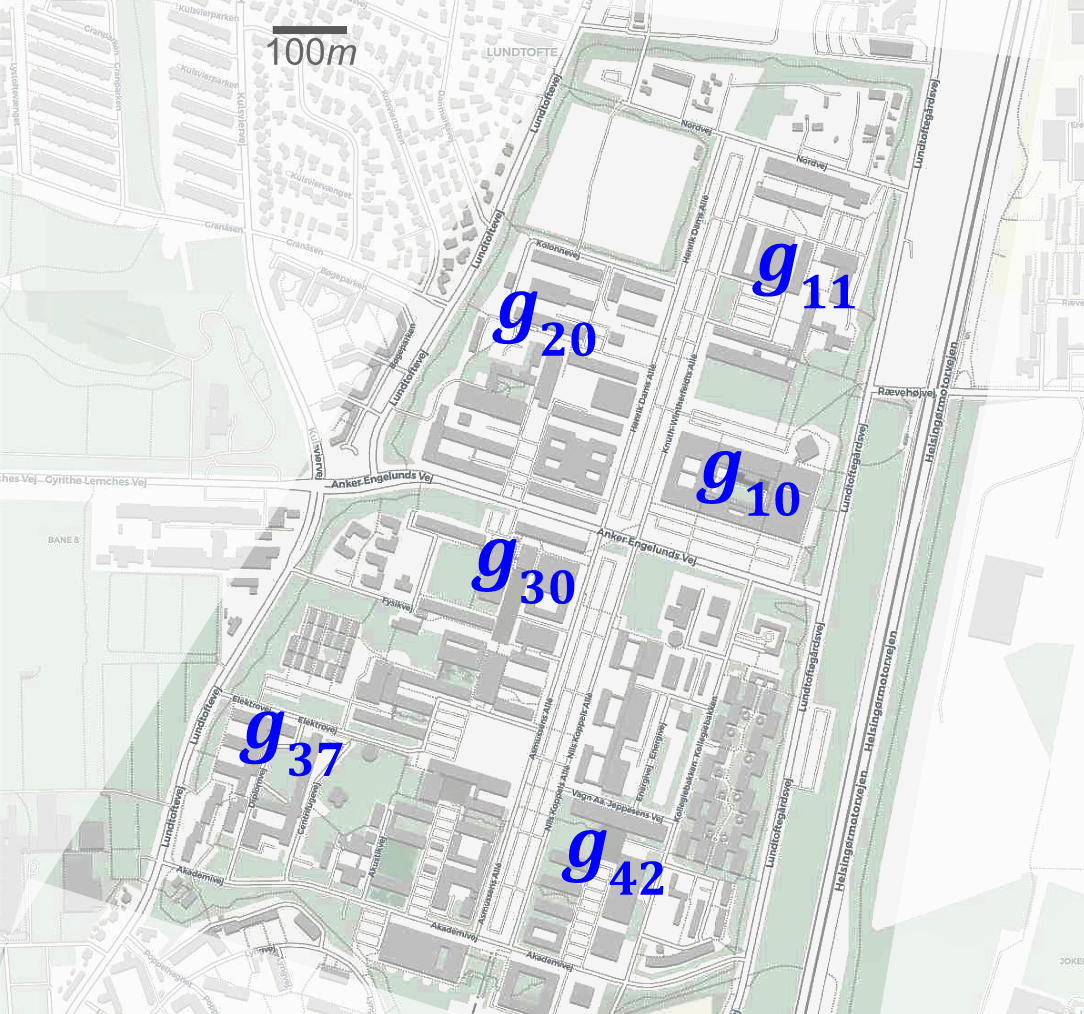}
        \caption{} \label{fig:dtu}
    \end{subfigure}
    \hfill
    \begin{subfigure}[b]{0.71\textwidth}
        \includegraphics[trim={0 0.5cm 0 0}, clip, width=\textwidth]{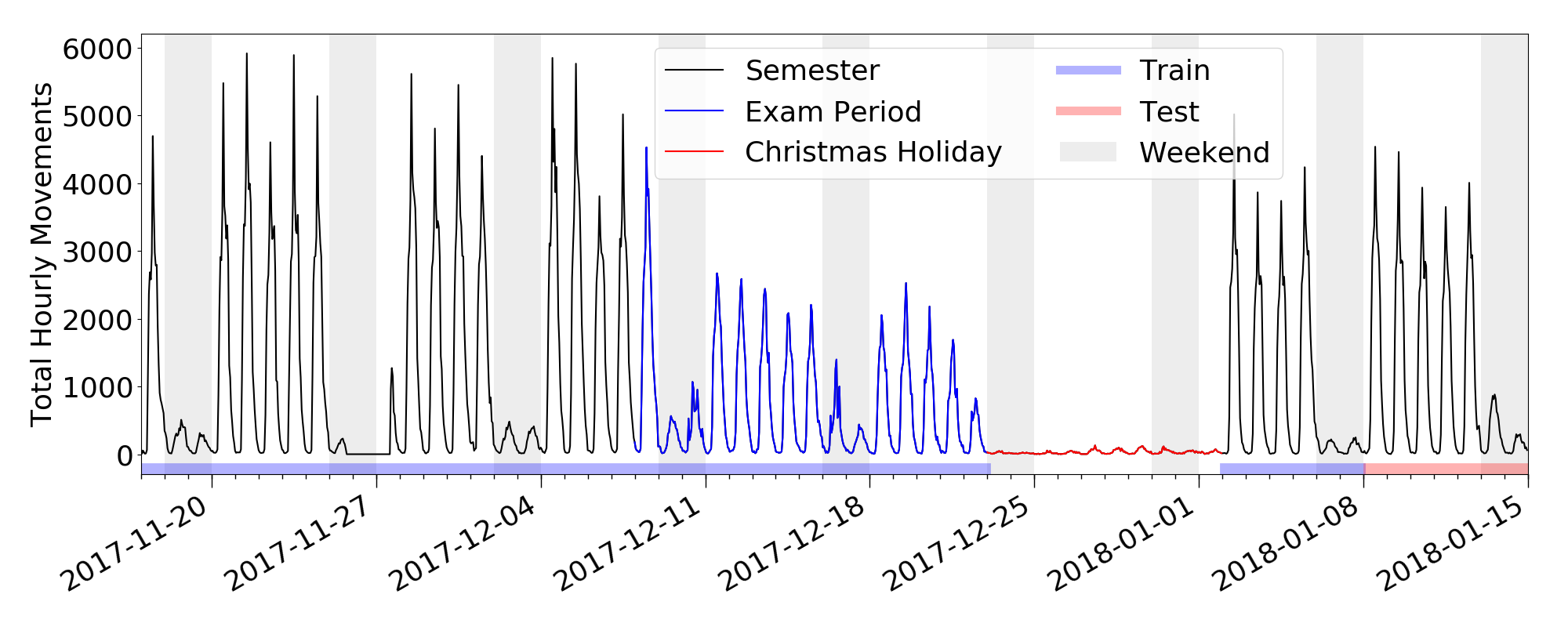}
        \caption{} \label{fig:total_ts}
    \end{subfigure}
    \caption{Data in our case study: (\ref{fig:dtu}) Observed locations in DTU Lyngby Campus, each covering one or more buildings. (\ref{fig:total_ts}) Observed hourly movements between all locations, with dates specified for Mondays.}
\end{figure}

The data in this case study consists of crowd movements between $6$ different locations in the Lyngby campus of the Technical University of Denmark (DTU), illustrated in \fgrref{fig:dtu}.
We label these locations as $g_{10}, g_{11}, g_{20}, g_{30}, g_{37}, g_{42}$ per internal building numbering in DTU, hence label numbers are not necessarily consecutive.  
In each location, WiFi access points probe for wireless devices, and we are given hourly aggregated counts of wireless devices that change location.
In total, we have $6 \cdot 6 - 6 = 30$ time series of hourly aggregated movements, one for each OD pair, ranging from 17-Nov-2017 00:00 to 14-Jan-2018 23:00.

The aggregated counts contain no information about individual devices.
We use the counts of wireless devices as a proxy for counts of people on campus, and assume that approximation errors amount to systematic noise in the observations.
For example, we assume that each person carries a personally fixed number of wireless devices (phone, laptop, tablet, etc.), and thus increases the approximated total by a consistent overhead. 

\subsection{Quantile Regression} \label{sec:qr}

As explained in Section \ref{sec:overview}, the total number of movements from each Origin to each Destination at each lag follows a latent marginal distribution. 
The density of this distribution can be estimated through quantiles of its Cumulative Distribution Function (CDF), as follows. 

\begin{figure}[ht!]
    \centering
    \begin{subfigure}{\textwidth}
    \centering
    \includegraphics[trim={4.3cm, 0cm, 5cm, 0cm}, clip, width=\textwidth]{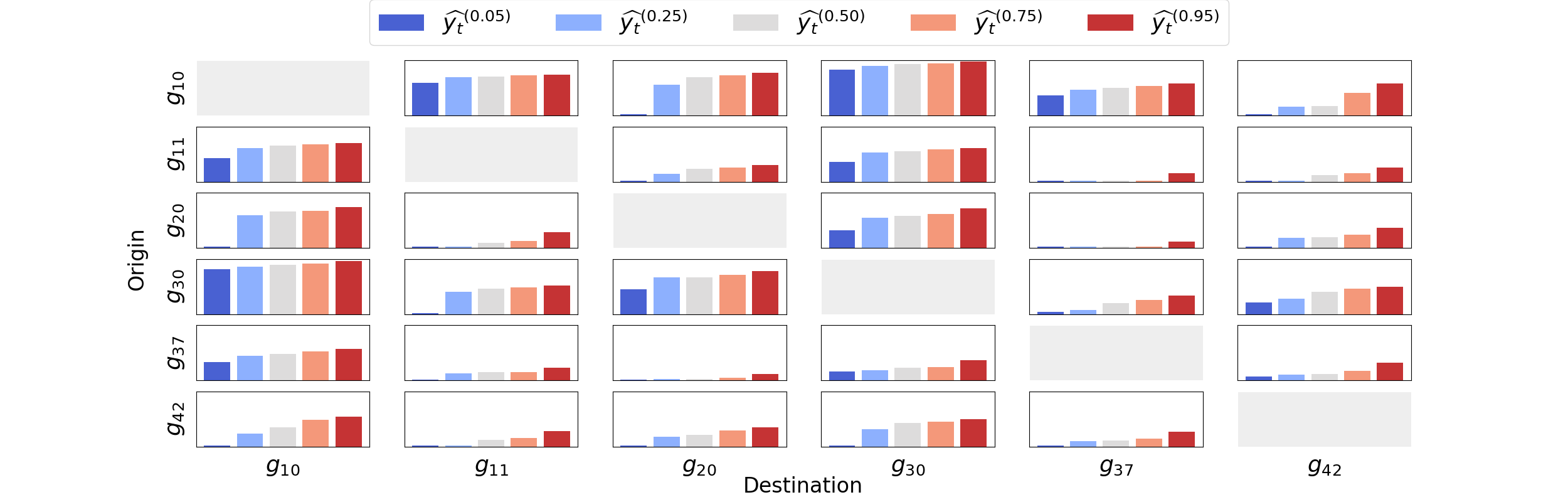}
    \end{subfigure}
    \begin{subfigure}{\textwidth}
    \centering
    \includegraphics[width=\textwidth, trim={4.3cm, 0cm, 5cm, 1cm}, clip]{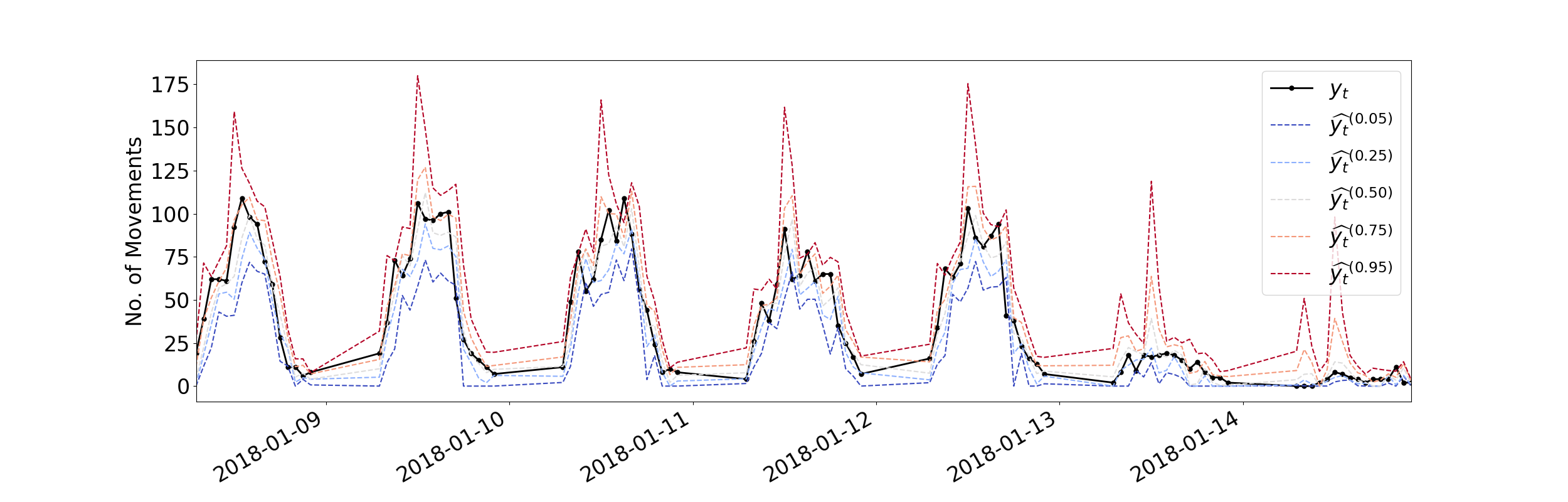}
    \end{subfigure}
    \caption{Example output of a Quantile Regression model ($LQR_3^{Ind}$). Top: estimated quantiles for every OD pair at one lag ($t = $ 2018-01-08 20:00), where $y$ axes are logarithmic between $0$ and $200$. Bottom: estimated quantiles for the entire test set of one OD pair ($g_{42} \rightarrow g_{10}$).}
    \label{fig:example_qr_output}
\end{figure}

For any OD pair and lag $t$, let $y_t$ denote the observed number of movements from origin to destination at lag $t$.
$y_t$ is thus a realization from a random variable $Y_t$, which has the corresponding latent marginal distribution of total movements.
Next, for any $0 < q < 1$, let $y_t^{(q)}$ denote the $q$'th quantile of the CDF of $Y_t$, i.e., the smallest real that the CDF maps to $q$.
We shall estimate $y_t^{(q)}$ for each $q \in Q = \{5\%, 25\%, 50\%, 75\%, 95\%\}$, and so obtain the following approximation of the marginal distribution:
\begin{equation}
\label{eq:preddist}
\Pr \left( 0 \leq y_t \leq \predhat{y}_t^{(0.05)} \right),
\Pr \left( \predhat{y}_t^{(0.05)} < y_t \leq \predhat{y}_t^{(0.25)} \right),
\dots,
\Pr \left( \predhat{y}_t^{(0.75)} < y_t \leq \predhat{y}_t^{(0.95)} \right),
\Pr \left( \predhat{y}_t^{(0.95)} < y_t \right)
\end{equation}
where $\predhat{y}_t^{(q)}$ is our estimation of the true $y_t^{(q)}$.
This manner of density estimation is thus named "Quantile Regression" (QR). To illustrate, \fgrref{fig:example_qr_output} shows an example of $\predhat{y}_t^{(q)}$ for one of our QR models, which we define later.

Before modeling, we need to treat a couple of issues in the time series, which might adversely affect model performance.
First, for some of the OD pairs, the time series of observations is non-stationary. 
Second, for all OD pairs, there are nearly no movements on campus during 23:00$\dots$06:59 and during Christmas holiday, $\left[\text{23-Dec-2017}, \dots, \text{1-Jan-2018}\right]$. 
We thus transform the time series of each OD pair in two steps.
First, we difference the time series as $\{y'_t\} = \{y_t - y_{t - 1}\}$, which is stationary for all OD pairs, as we verify through augmented Dickey-Fuller test with $p\text{-value} = 1\%$. 
Second, we remove from $\{y'_t\}$ all lags in hours 23:00$\dots$06:59 (these can also be used for shuttle maintenance) and in the Christmas holiday.

We are now ready to fit and test various demand prediction models on the remaining lags in the transformed time series, which we partition as follows: train set $T_{train}$ consists of all remaining lags in $\left[\text{17-Nov-2017}, \dots, \text{7-Jan-2018}\right]$ ($52$ days), whereas test set $T_{test}$ consists of all other remaining lags in $\left[\text{8-Jan-2018}, \dots, \text{14-Jan-2018}\right]$ ($7$ days), as illustrated in \fgrref{fig:total_ts}.
We train each QR model on $T_{train}$, and then evaluate it on $T_{test}$ by calculating the following measures for each OD pair:

\begin{itemize}

\item Mean Titled Loss:
\begin{equation*}
\mtl = \sum_{q \in Q} \frac{1}{\left| T_{test} \right|} \sum_{t \in T_{test}} \max{\{
    q (y_t - \predhat{y}_t^{(q)}),
    \left(q - 1\right) (y_t - \predhat{y}_t^{(q)})
    \}
}
\end{equation*}

\item Percent of measured values that fall within the $5\%-95\%$ quantile range:
\begin{equation*}
\icp = \frac{1}{\left| T_{test} \right|} \left| \left\{t \in T_{test} \mid \predhat{y}_t^{(0.05)} \leq y_t \leq \predhat{y}_t^{(0.95)} \right\}\right|
\end{equation*}

\item Mean length of $5\%-95\%$ quantile range: 
\begin{equation*}
\mil = \frac{1}{\left| T_{test} \right|} \sum_{t \in T_{test}} \predhat{y}_t^{(0.95)} - \predhat{y}_t^{(0.05)}
\end{equation*}

\item Number of pairwise quantile crossings:
\begin{equation*}
    \text{\#cross} = \sum_{t \in T_{test}} \left| \left\{ \left( q_i \in Q, q_j \in Q \right) \mid q_i < q_j \wedge \predhat{y}_t^{(q_i)} > \predhat{y}_t^{(q_j)} \right\} \right|
\end{equation*}
\end{itemize}

Because $\mtl$ is a weighted mean of prediction errors, a QR prediction model is considered better if it achieves lower total $\mtl$ over all OD pairs.
The other measures are averaged for each model over all OD pairs, and we use them as additional indicators of performance quality: it is preferred to bring $\icp$ closer to $95\% - 5\% = 90\%$ while simultaneously obtaining lower mean $\mil$, and it is also preferred to obtain fewer quantile crossings.

For the remainder of this Section, we define and experiment with the following types of Quantile Regression models: Historical Percentiles, Linear QR, Deep QR, and Gradient Boosting QR. 
\tblref{tab:qr_performance} summarizes the overall predictive performance of all models, and highlights the best performing model.

\begin{table*}[ht]
    \footnotesize
    \caption{Predictive performance of Quantile Regression models.}
    \label{tab:qr_performance}
    \centering
    \begin{tabular}{l l l l l} 
    \hline
    \thead{Model} & \thead{Total\\$\mtl$} & \thead{Mean\\$\icp$} & \thead{Mean\\$\mil$} & \thead{Mean\\\#cross} \\
    \hline
    $HP^{Ind}$ & $446.071$ & $0.731 \ (\pm 0.051)$ & $26.129 \ (\pm 32.929)$ & $0.000 \ (\pm 0.000)$ \\
    \hline
    $LQR_1^{Ind}$ w. Seasonality & $437.805$ & $0.879 \ (\pm 0.055)$ & $38.741 \ (\pm 48.215)$ & $13.800 \ (\pm 20.908)$ \\
    $LQR_2^{Ind}$ w/o Seasonality & 459.428 & $0.777 \ (\pm 0.055)$ & $31.287 \ (\pm 39.255)$ & $15.667 \ (\pm 9.321)$ \\
    $LQR_3^{Ind}$ w. Sorting & $437.218$ & $0.882 \ (\pm 0.053)$ & $38.781 \ (\pm 48.265)$ & $0.000 \ (\pm 0.000)$ \\
    $LQR_4^{Ind}$ w. Exams & $437.235$ & $0.900 \ (\pm 0.049)$ & $39.692 \ (\pm 49.696)$ & $0.000 \ (\pm 0.000)$ \\
    $LQR_5^{Ind}$ w/o first week & $444.537$ & $0.896 \ (\pm 0.045)$ & $40.463 \ (\pm 50.574)$ & $0.000 \ (\pm 0.000)$\\
    \hline
    $DNN^{Ind}$ FC Linear, Common LR & $475.826$ & $0.925 \ (\pm 0.041)$ & $42.043 \ (\pm 55.191)$ & $0.000 \ (\pm 0.000)$\\
    $DNN^{Ind}$ FC Linear, LR per OD & $474.174$ & $0.930 \ (\pm 0.035)$ & $41.384 \ (\pm 50.659)$ & $0.000 \ (\pm 0.000)$\\
    $DNN^{Ind}$ FC Linear, LR per OD and $q$ & $477.845$ & $0.949 \ (\pm 0.034)$ & $52.220 \ (\pm 69.943)$ & $0.000 \ (\pm 0.000)$\\
    \hline
    $GBoost^{Ind}$, Common Params & $455.074$ & $0.839 \ (\pm 0.042)$ & $40.041 \ (\pm 56.008)$ & $0.000 \ (\pm 0.000)$\\
    $GBoost^{Ind}$, Params per OD & $455.295$ & $0.838 \ (\pm 0.050)$ & $39.037 \ (\pm 55.906)$ & $0.000 \ (\pm 0.000)$\\ 
    $GBoost^{Ind}$, Params per OD and $q$ & $446.999$ & $0.846 \ (\pm 0.041)$ & $39.508 \ (\pm 56.020)$ & $0.000 \ (\pm 0.000)$\\
    \hline \hline
    $LQR^{Mul}_1$ All ODs Together & $494.227$ & $0.918 \ (\pm 0.104)$ & $41.663 \ (\pm 48.100)$ & $0.000 \ (\pm 0.000)$\\
    $LQR^{Mul}_2$ similar to $VAR(1)$ & $485.963$ & $0.876 \ (\pm 0.046)$ & $39.421 \ (\pm 50.775)$ & $0.000 \ (\pm 0.000)$\\
    \hline
    $DNN_1^{Mul}$ All Quantiles Together & $468.709$ & $0.907 \ (\pm 0.047)$ & $36.431 \ (\pm 43.771)$ & $0.000 \ (\pm 0.000)$\\
    $DNN_2^{Mul}$ Parameter Sharing & $777.361$ & $0.852 \ (\pm 0.072)$ & $45.634 \ (\pm 66.715)$ & $0.000 \ (\pm 0.000)$\\
    \hline
    \rowcolor{yellow!27} 
    $GBoost_1^{Mul}$ Common Params & $433.980$ & $0.883 \ (\pm 0.040)$ & $34.621 \ (\pm 39.883)$ & $0.000 \ (\pm 0.000)$\\
    $GBoost_2^{Mul}$ Params per $q$ & $437.528$ & $0.884 \ (\pm 0.046)$ & $37.649 \ (\pm 43.113)$ & $0.000 \ (\pm 0.000)$\\
    \hline
    \end{tabular}
\end{table*}

\subsection{Independent Models}

We begin by fitting models independently for each $q \in Q$ and each OD pair.
For each model type, we thus fit a total of $5 \cdot 30 = 150$ independent models, and evaluate their overall performance as explained above.
We use superscript $^{Ind}$ as part of a model name (e.g., as in $HP^{Ind}$) to indicate such independent fitting.

\subsubsection{Modeling by Historical Percentiles}

For any lag $t$, let $TOD(t) \in \{7, \dots, 22\}$ denote the time-of-day of $t$, and let $DOW(t) \in \{0, \dots, 6 \}$ denote the day-of-week of $t$, so that $0 \equiv \text{Monday}$. 
Our first model is a naive Historical Percentiles model $HP^{Ind}$, in which for each OD pair and each $q \in Q$, $\predhat{y}_t^{(q)}$ is the $q$'th percentile of $$
\{ y_k \mid k < t \wedge DOW(k) = DOW(t) \wedge TOD(k) = TOD(t) \}
$$

As Table $\ref{tab:qr_performance}$ shows, $HP^{Ind}$ achieves $\icp$ far from $90\%$, which indicates that the underlying time series is not a simple repetition of historical patterns. 
Moreover, if instead of $Q$ we fit $HP^{Ind}$ on the $0\%$'th and $100\%$'th quantiles -- namely the minima and maxima in the train set -- we obtain mean $ICP$ only $80\%\ (\pm 5.5\%)$, which is far from $100\%$. 
This means that for some OD pairs, there are movement counts in the test set that lie outside of the range of values in the train set, which confirms that historical percentiles alone are insufficient predictors for this data.

\subsubsection{Linear Quantile Regression}

We proceed to more flexible Linear Quantile Regression (LQR) models, where the estimated number of movements is modeled as follows:
\begin{align*}
\predhat{y}_t^{(q)} &= \beta^{TOD}_7 d_t^{(7)} + \dots + \beta^{TOD}_{22} d_t^{(22)} \\
&+ \beta^{DOW}_0 w_t^{(0)} + \dots + \beta^{DOW}_6 w_t^{(6)} \\
&+ \beta_{-1} y_{t - 1} + \dots + \beta_{-24} y_{t - 24} \numberthis \label{eq:lqr}
\end{align*}
where $d_t^{(i)} \in \{0, 1\}$ is $1$ iff $i = TOD(t)$, $w_t^{(i)} \in \{0, 1\}$ is $1$ iff $i = DOW(t)$, and the parameters to be estimated are $\beta$'s (with subscripts corresponding to $TOD$, $DOW$, and lag).
As for all other QR prediction models, the loss function to be minimized is the total $MTL$.
We train all LQR models via Iterative Weighted Least Squares with an Epanechnikov kernel and Hall-Sheather bandwidth selection \cite{hall1988distribution}, and clip any negative predictions to zero. 

Our first linear model $LQR_1^{Ind}$ achieves better total $MTL$ and significantly better mean $\icp$ than $HP^{Ind}$.
Next, we attempt to further improve performance by removing seasonality from the data.
That is, we transform ${y'_t}$ into $y''_t = \left( y'_t - \overline{y'}_t \right) / \widetilde{y'}_t$, where $\overline{y'}_t$ and $\widetilde{y'_t}$ are, respectively, the mean and standard deviation of $\{y'_k \mid k \in T_{train} \wedge DOW(k) = DOW(t) \wedge TOD(k) = TOD(t) \}$. 
However, $LQR_2^{Ind}$ performs worse on $\{y''_t\}$ than $LQR_1^{Ind}$ does on $\{y'_t\}$.

We thus proceed to run LQR again on $\{y'_t\}$, and this time eliminate quantile crossings by sorting $\predhat{y}_t^{(0.05)}, \dots, \predhat{y}_t^{(0.95)}$ in ascending order for every $t$. 
This results in a better performing model $LQR_3^{Ind}$, for which we have illustrated some predictions earlier in \fgrref{fig:example_qr_output}.

During $\left[ \text{8-Dec-2017}, \dots, \text{22-Dec-2017} \right]$, exams took place on campus.
Hence we next add to each data vector a binary feature $m_t$, which indicates whether the data is in the exam period, namely:
\begin{align*}
\predhat{y}_t^{(q)} &= \beta^{TOD}_7 d_t^{(7)} + \dots + \beta^{TOD}_{22} d_t^{(22)} \\
&+ \beta^{DOW}_0 w_t^{(0)} + \dots + \beta^{DOW}_6 w_t^{(6)} \\
&+ \beta^{EX} m_t \\
&+ \beta_{-1} y_{t - 1} + \dots + \beta_{-24} y_{t - 24} \numberthis \label{eq:lqr_with_exam}
\end{align*}
The corresponding model is $LQR_4^{Ind}$, which we train with quantile sorting.
$LQR_4^{Ind}$ yields nearly the same total $MTL$ as does $LQR_3^{Ind}$, and mean $\icp$ noticeably closer to $90\%$, with a small increase in mean $\mil$. 
We thus designate $LQR_4^{Ind}$ as the best performing model so far.

We next proceed to Deep Quantile Regression, where the training process requires us to reserve a week of the data for validation.
Hence for fair comparison with LQR, we also build and estimate model $LQR_5^{Ind}$, where we omit the first week from the train set.
Consequently, $LQR_5^{Ind}$ performs worse than $LQR_4^{Ind}$, where the train set includes the first week. 

\subsubsection{Deep Quantile Regression}

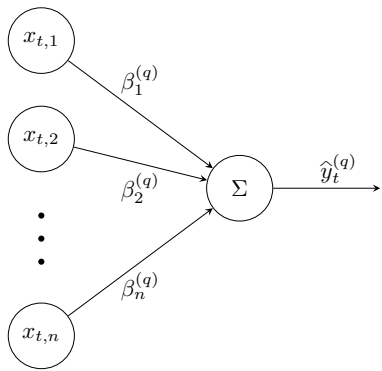
\begin{figure}[ht]
    \centering
    \scalebox{0.87}{
    \begin{tikzpicture}[x=1.5cm, y=1.5cm, >=stealth]
    \node [align=center, above] at (0, -1.5) {}; 
    \node [align=center, above] at (2, -1.5) {}; 
    \node [neuron] (input0) at (0, -2) {$x_{t,1}$};
    \node [neuron] (input1) at (0, -3) {$x_{t,2}$};
    \node [missing]  (missing1) at (0, -4) {};
    \node [neuron] (inputN) at (0, -5) {$x_{t,n}$};
    \node [neuron] (sigma0) at (2, -3.5) {$\Sigma$}
        edge [<-] (input0)
        edge [<-] (input1) 
        edge [<-] (inputN);
    \node [] (unseen0) at (3.5, -3.5) {}
        edge [<-] (sigma0) node at (3, -3.3) {$\predhat{y}_t^{(q)}$};
    \node (beta1) at (1, -2.4) {$\beta^{(q)}_1$};
    \node (beta2) at (1, -3.5) {$\beta^{(q)}_2$};
    \node (betaN) at (1, -4.5) {$\beta^{(q)}_n$};
    \end{tikzpicture}
    }
    \caption{$DNN^{Ind}$, where $x_{t,1}, \dots, x_{t,n}$ are input features at lag $t$, and $\beta^{(q)}_1, \dots, \beta^{(q)}_n$ are trainable parameters.}
    \label{fig:dnn_ind}    
\end{figure}


Whereas LQR models are parametric and linear, Deep Neural Network (DNN) models provide a non-parametric approximation of the true predictive distribution.
As a baseline, we wish to first obtain similar performance for $DNN$ as for $LQR$, hence our first deep QR model $DNN^{Ind}$ is as in \fgrref{fig:dnn_ind}: a fully-connected, feed-forward neural network, where a single linear unit combines all input features.
While training $DNN^{Ind}$, we use the first week of $T_{train}$ as a validation set for early stop.
We thus aim for $DNN^{Ind}$ to first perform as well as $LQR_5^{Ind}$, where we have omitted the first week from the train set too.

The performance of DNN strongly depends on hyper-parameter selection, hence before training, we tune a learning rate in $[10^{-8}, 1]$ using Bayesian Optimization: a common approach for hyper-parameter tuning \cite{snoek2012practical}.
To this end, we partition $T_{train}$ into three subsets: the first $7$ days in $T_{opt}^{test}$ for testing, the following $7$ days in $T_{opt}^{val}$ for validation, and the remaining days in $T_{opt}^{train}$ for training.
In every iteration, the optimizer tries a different learning rate to train the DNN on $T_{opt}^{train}$, while using $T_{opt}^{val}$ for early stop, and then obtains the total $MTL$ of the trained DNN on $T_{opt}^{test}$.
The optimal learning rate is the one which minimizes this total $MTL$ after at most $100$ iterations.

We note again that learning rate optimization is done separately of and prior to actually fitting the parameters of the DNN models.
That is, we use data sets $T_{opt}^{val}, T_{opt}^{train}, T_{opt}^{test}$ only during learning rate optimization, whereas during actual model fitting, we use $T_{train}$ and $T_{test}$ as for all other demand models (while internally using the first week of $T_{train}$ to avoid overfitting). 
We also note that while the same learning rate may be common to several models, we still fit every $DNN^{Ind}$ model independently for each $q \in Q$ and each OD pair.

\begin{figure}[ht]
    \centering
    \begin{subfigure}[b]{0.49\linewidth}
    \centering
    \includegraphics[width=\linewidth]{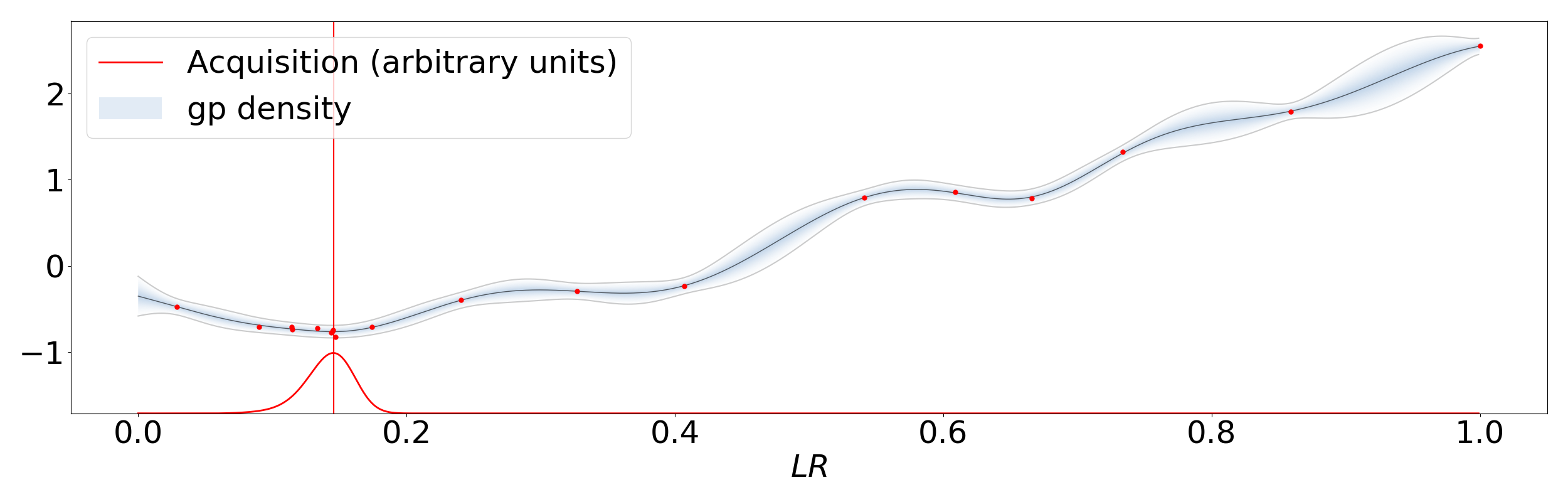}
    \caption{Common LR}
    \label{fig:dnn_lr_common}
    \end{subfigure}
    \hfill
    \begin{subfigure}[b]{0.49\linewidth}
    \centering
    \includegraphics[width=\linewidth]{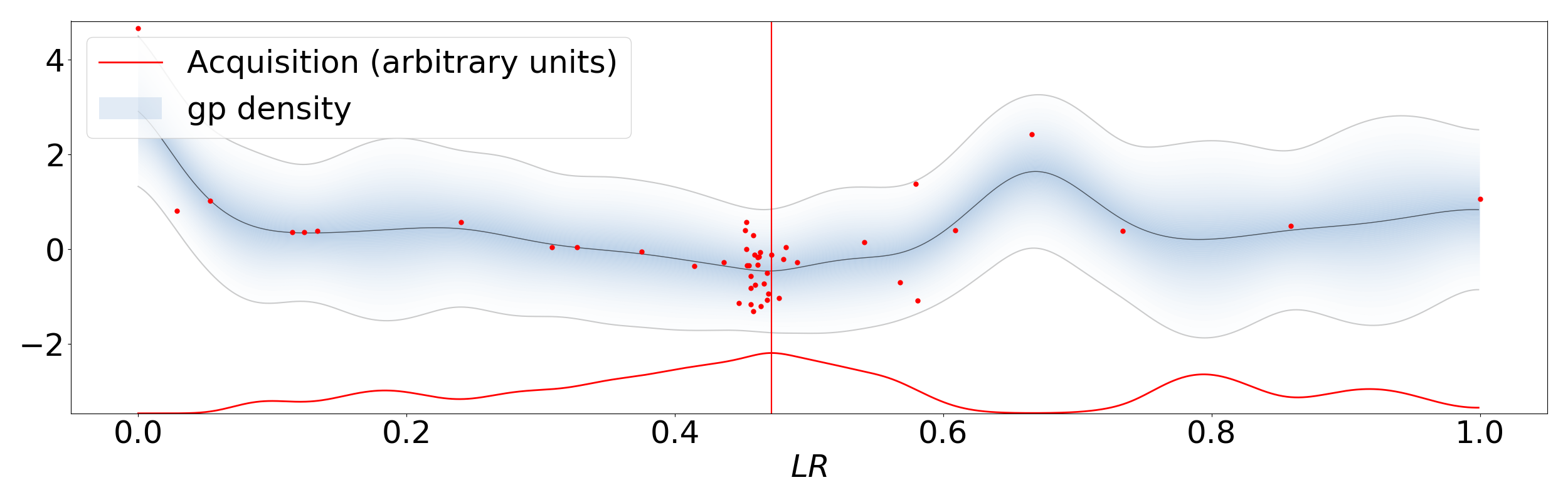}
    \caption{LR per OD ($g_{20} \rightarrow g_{42}$)}
    \label{fig:dnn_lr_per_od}
    \end{subfigure}
    \hfill
    \begin{subfigure}[b]{0.49\linewidth}
    \centering
    \includegraphics[width=\linewidth]{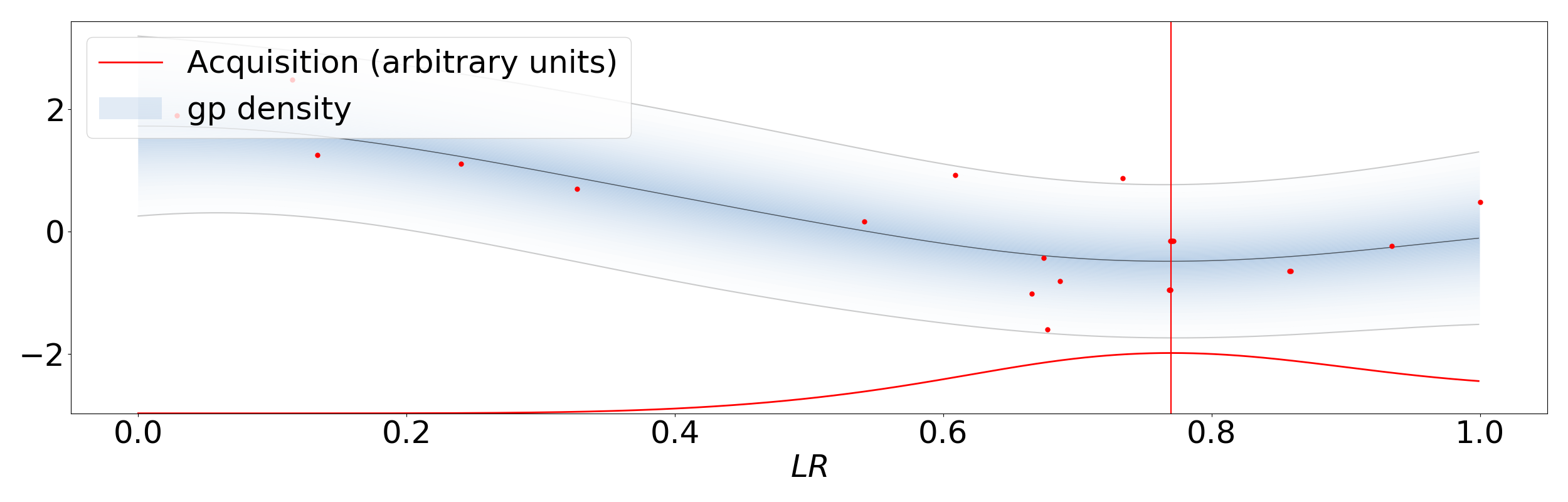}
    \caption{LR per OD and $q$ ($g_{20} \rightarrow g_{42}, q=0.95$)}
    \label{fig:dnn_lr_per_od_and_q}
    \end{subfigure}
    \caption{Bayesian Optimization of learning rate (LR) of $DNN^{Ind}$. The best LR, indicated by a vertical line, can shift considerably for the same OD pair under different optimization resolutions.} 
    \label{fig:bopt_dnn}
\end{figure}

\fgrref{fig:bopt_dnn} describes the learning rate optimization for independent Deep QR models. First, we optimize a single learning rate, common to all OD pairs and all $q \in Q$ (\fgrref{fig:dnn_lr_common}).
The common learning rate yields worse performance than $LQR_5^{Ind}$, as shown in \tblref{tab:qr_performance}. 
We thus next optimize a learning rate independently for each OD pair (\fgrref{fig:dnn_lr_per_od}).
This results in a little improvement over the common learning rate, but performance is still noticeably worse than $LQR_5^{Ind}$. Finally, we optimize the learning rate independently for every combination of OD and $q \in Q$ (\fgrref{fig:dnn_lr_per_od_and_q}), but gain no performance improvement.

In summation, Deep Quantile Regression did not perform as well as Linear Quantile Regression, despite learning rate optimization.
The dataset is thus not large enough for effective deep learning with backpropagation, hence we do not attempt to further improve performance, e.g., using kernel regularization, dropout, or recurrent units. 
As an alternative attempt at non-parametric QR, we next experiment with Gradient Boosting.

\subsubsection{Gradient Boosting} \label{sec:gboost}

\begin{figure}[htb!]
    \centering
    \scalebox{0.78}{
    \begin{tikzpicture}[x=1.5cm, y=1.5cm, >=stealth]
    \draw[color=gray] (-0.5, 0.3) -- (2.1, 0.3) -- (2.1, -4) -- (-0.5, -4) -- (-0.5, 0.3);
    \node [sqneuron] (input10) at (0, -2) {};
    \node [sqneuron] (input11) at (1.5, -3) {$14.3$}
        edge [<-] (input10);
    \node [sqneuron] (input12) at (1.5, -1) {$8.94$}
        edge [<-] (input10);
    \node (geq11) at (0.4, -1.3) {$x_{t,3} \geq 0.2$};
    \node (geq12) at (0.4, -2.8) {$x_{t,3} < 0.2$};
    
    \node [cdots] (dts) at (2.65, -1.8) {};
    
    \draw[color=gray] (3.1, 0.3) -- (7.7, 0.3) -- (7.7, -4) -- (3.1, -4) -- (3.1, 0.3);
    \node [sqneuron] (input21) at (3.7, -2) {};
    \node [sqneuron] (input22) at (5.2, -3) {}
        edge [<-] (input21);
    \node [sqneuron] (input23) at (5.2, -1) {}
        edge [<-] (input21);
    \node (geq21) at (4., -1.3) {$x_{t, 1} \leq -0.56$};
    \node (geq22) at (4., -2.8) {$x_{t, 1} > -0.56$};
    \node [sqneuron] (input25) at (7.2, -0.5) {$1.4$}
        edge [<-] (input23);
    \node (geq23) at (6., -1.6) {$x_{t, n} \leq     -0.11$};
    \node (geq24) at (6., -0.5) {$x_{t, n} > -0.11$};
    \node [sqneuron] (input24) at (7.2, -1.6) {$0.88$}
        edge [<-] (input23);
    \node [sqneuron] (input26) at (7.2, -2.5) {$1$}
        edge [<-] (input22);
    \node [sqneuron] (input27) at (7.2, -3.6) {$2.1$}
        edge [<-] (input22);
    \node (geq25) at (6.1, -3.6) {$x_{t, 6} > 0.02$};
    \node (geq26) at (6.1, -2.5) {$x_{t, 6} \leq 0.02$};
    \node (t1) at (-0.2, 0) {$T_1$};
    \node (t1) at (3.5, 0) {$T_{68}$};
    \end{tikzpicture}
    }
    \caption{A Gradient Boosting model passes the input $x_t = \left(x_{t,1}, \dots, x_{t,n}\right)$ through each decision tree until reaching a leaf, then outputs the sum of all these leaves. Edge labels correspond to decision rules, which we arbitrarily instantiate here for demonstration purposes.}
    \label{fig:gboost}
\end{figure}
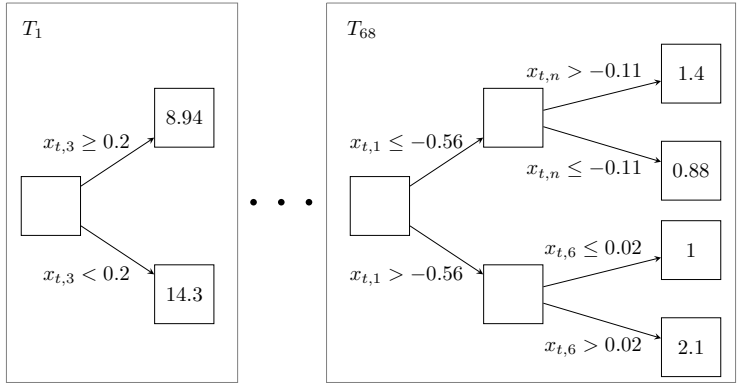

Gradient Boosting (GBoost) \cite{friedman2002stochastic} builds an ensemble of regression trees incrementally using gradient descent. 
As in previous prediction models, the loss function being boosted is total $MTL$. 
\fgrref{fig:gboost} shows an example of $GBoost^{Ind}$ for one OD pair and a single $q \in Q$.

\begin{figure}[ht]
    \begin{subfigure}{0.49\linewidth}
    \centering
    \includegraphics[width=\linewidth]{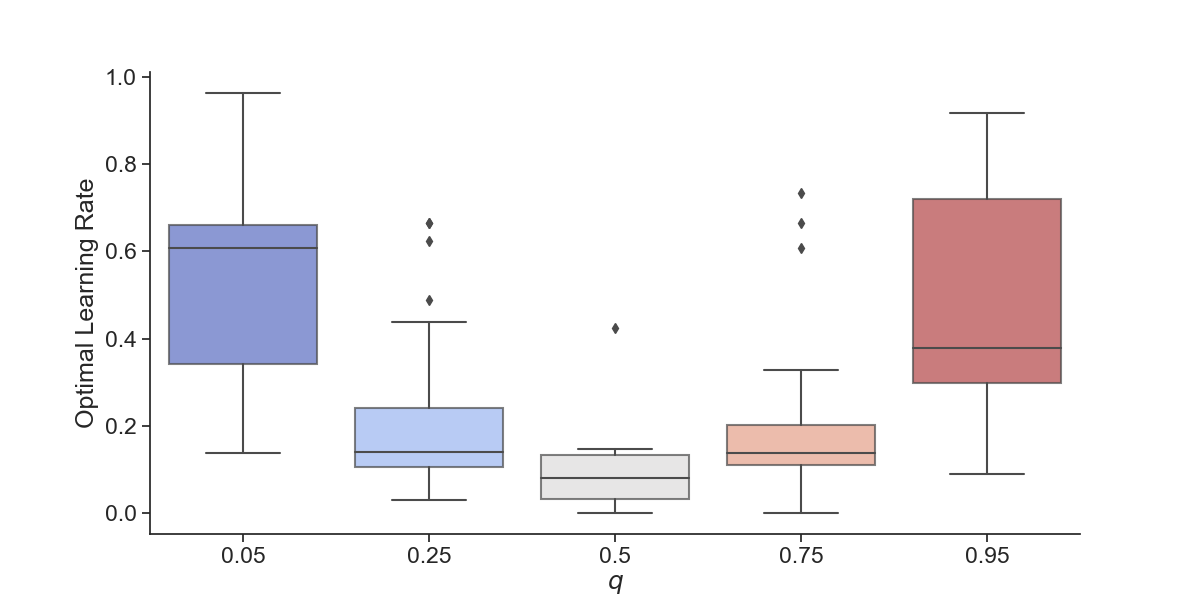}
    \caption{$DNN^{Ind}$}
    \label{fig:opt_lr_dnn}
    \end{subfigure}
    \hfill
    \begin{subfigure}{0.49\linewidth}
    \centering
    \includegraphics[width=\linewidth]{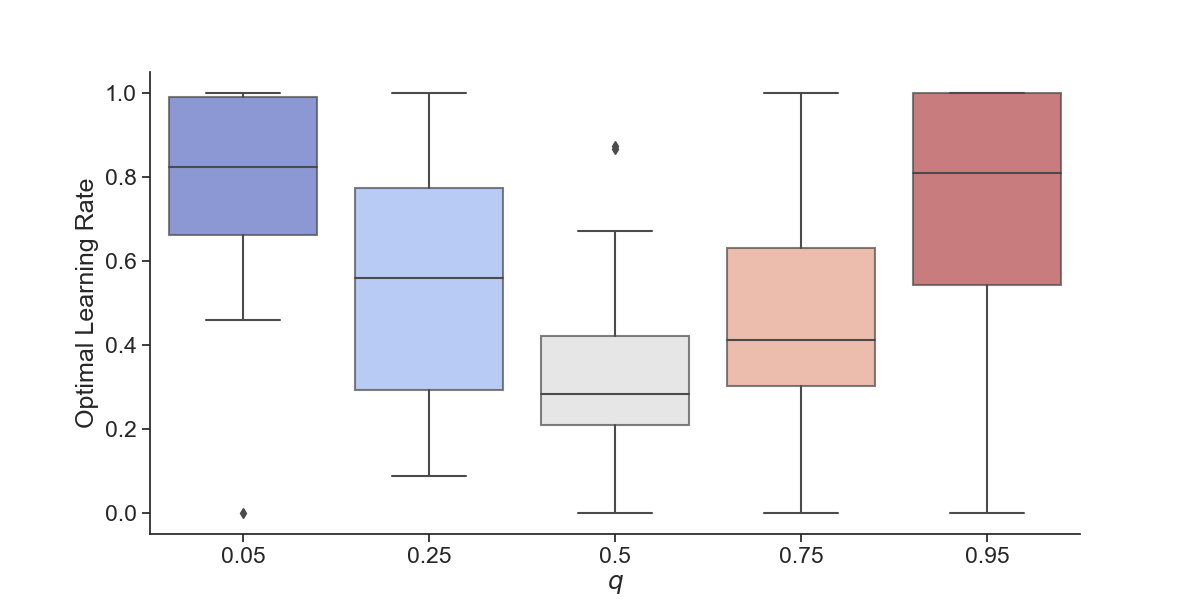}
    \caption{$GBoost$}
    \label{fig:opt_lr_gboost}
    \end{subfigure}    
    \caption{Box plot of distribution of optimal learning rate over all OD pairs, for each $q \in Q$.}
\end{figure}

As for $DNN^{Ind}$, we partition $T_{train}$ as $T_{opt}^{train}, T_{opt}^{val}, T_{opt}^{test}$, and use Bayesian Optimization to find the best hyper-parameters for $GBoost^{Ind}$: learning rate in $\left[10^{-8}, 1\right]$, maximum tree depth in $1..6$, and maximum number of trees in $1..200$.
\fgrref{fig:opt_lr_gboost} shows that for all $q \in Q$, the optimal learning for $GBoost^{Ind}$ follows a similar pattern of distributions as for $DNN^{Ind}$ in\fgrref{fig:opt_lr_dnn}, albeit with greater variance for $q \in \{0.25, 0.75\}$.
Nevertheless, \tblref{tab:qr_performance} shows that contrary to $DNN^{Ind}$, the performance of $GBoost^{Ind}$ improves when hyper-parameter optimization is carried out separately for each OD and $q \in Q$.
Compared to $LQR_5^{Ind}$, the best performing $GBoost^{Ind}$ has similar total $MTL$ and mean $\mil$, but worse mean $\icp$. 

\subsection{Modeling Spatio-Temporal Dependencies}

\begin{figure}[ht]
    \centering
    \includegraphics[width=0.6\linewidth]{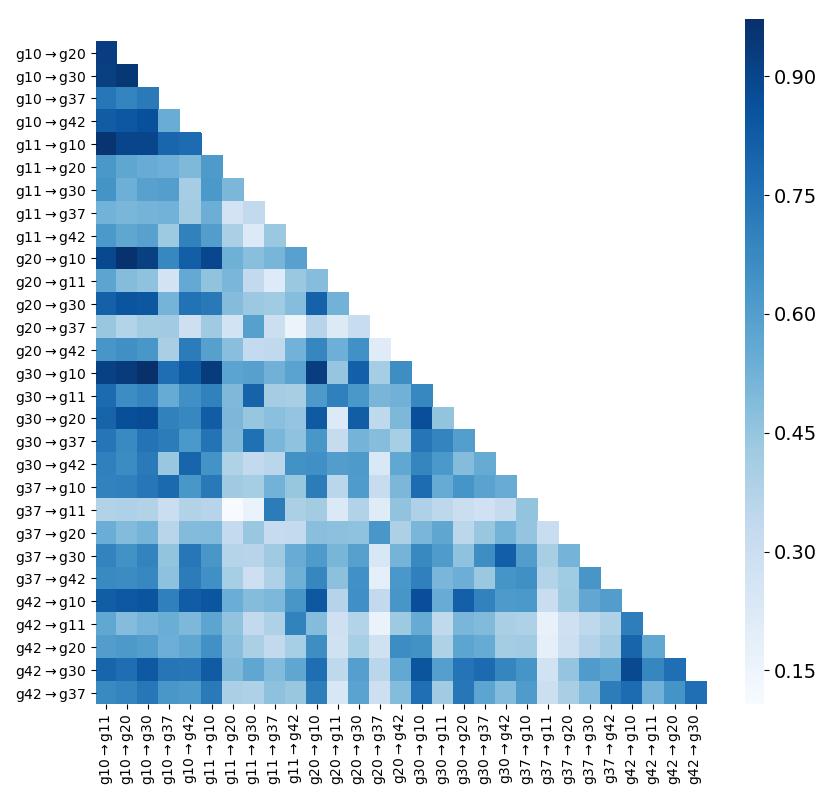}
    \caption{Pearson correlation between the time series of every couple of different OD pairs.}
    \label{fig:corr_od_pairs}
\end{figure}

For every type of prediction model above, we have built $150$ independent models, each pertaining to a different OD pair and $q \in Q$.
Let us now examine models which process together multiple OD pairs or multiple quantiles, and so may be able to exploit spatio-temporal dependencies in the data.
Indeed, Figure \ref{fig:corr_od_pairs} shows that for some couples of OD pairs, the time series of movements are strongly correlated.

\subsubsection{Linear Quantile Regression on Multiple Time Series}
We begin with a linear QR model $LQR_1^{Mul}$ which processes all $30$ OD pairs together, independently for each $q \in Q$, as:
\begin{align*}
\predhat{y}_t^{(q)} &= \beta^{TOD}_7 d_t^{(7)} + \dots + \beta^{TOD}_{22} d_t^{(22)} \\
&+ \beta^{DOW}_0 w_t^{(0)} + \dots + \beta^{DOW}_6 w_t^{(6)} \\
&+ \beta^{EX} m_t \\
&+ \beta_{-1} y_{t - 1} + \dots + \beta_{-24} y_{t - 24} \\
&+ \beta^{OD}_1 o_1 + \dots \beta^{OD}_{30} o_{30} \numberthis
\label{eq:lqr_1_mv}
\end{align*}
where for an arbitrary, pre-fixed ordering of the $30$ OD pairs, $o_i$ is binary and indicates whether the data corresponds to the $i$'th OD pair. 
We train and test $LQR_1^{Mul}$ using the same methods as for $LQR_4^{Ind}$. \tblref{tab:qr_performance} summarizes the performance of $LQR_1^{Mul}$, which is seen to be significantly worse than $LQR_4^{Ind}$.

Next, we try linear model $LQR_2^{Mul}$, where for each OD pair independently:
\begin{align*}
\predhat{y}_t^{(q)} &= \beta^{TOD}_7 d_t^{(7)} + \dots + \beta^{TOD}_{22} d_t^{(22)} \\
&+ \beta^{DOW}_0 w_t^{(0)} + \dots + \beta^{DOW}_6 w_t^{(6)} \\
&+ \beta^{EX} m_t \\
&+ \sum_{i=1}^{30} \sum_{k=1}^{p} \beta^{(i)}_{-k} y^{(i)}_{t - k} \numberthis
\end{align*}
where $p \in \mathbb{N}$ is a hyper-parameter, and $y^{(i)}_{t - k}$ is the $k$'th past lag of the $i$'th OD pair. 
$LQR_2^{Mul}$ is therefore similar to $VAR(p)$, namely Vector Autoregression of order $p$, where each of several response variables is modeled on the last $p$ lags of all response variables. 
We have experimented with $p=1, 2, 3, 4$, and noticed that performance deteriorated as $p$ increased. 
\tblref{tab:qr_performance} summarizes the performance of $LQR_2^{Mul}$ for $p=1$, which is seen to perform better than $LQR_1^{Mul}$, but still worse than $LQR_4^{Ind}$.

\subsubsection{Multivariate Deep Quantile Regression}

\begin{figure}[ht]
    \centering
    \begin{subfigure}{0.45\linewidth}
    \scalebox{0.8}{
    \begin{tikzpicture}[x=1.5cm, y=1.5cm, >=stealth]

    \node [align=center, above] at (0, -1) {}; 
    \node [align=center, above] at (2, -1) {}; 
    
    \node [neuron] (input0) at (0, -2) {$x_{t,1}$};
    \node [neuron] (input1) at (0, -3) {$x_{t,2}$};
    \node [missing]  (missing1) at (0, -4) {};
    \node [neuron] (inputN) at (0, -5) {$x_{t,n}$};
    
    \node [neuron] (sigma0) at (2, -1.5) {$\Sigma$}
        edge [<-] (input0) 
        edge [<-] (input1) 
        edge [<-] (inputN);
    \node [neuron] (sigma1) at (2, -2.5) {$\Sigma$}
        edge [<-] (input0) 
        edge [<-] (input1) 
        edge [<-] (inputN);
    \node [neuron] (sigma2) at (2, -3.5) {$\Sigma$}
        edge [<-] (input0) 
        edge [<-] (input1) 
        edge [<-] (inputN);
    \node [neuron] (sigma3) at (2, -4.5) {$\Sigma$}
        edge [<-] (input0) 
        edge [<-] (input1) 
        edge [<-] (inputN);
    \node [neuron] (sigma4) at (2, -5.5) {$\Sigma$}
        edge [<-] (input0) 
        edge [<-] (input1) 
        edge [<-] (inputN);
    
    \node [] (unseen0) at (3.5, -1.5) {}
        edge [<-] (sigma0) node at (3, -1.3) {$\predhat{y}_t^{(0.05)}$};
    \node [] (unseen1) at (3.5, -2.5) {}
        edge [<-] (sigma1) node at (3, -2.3) {$\predhat{y}_t^{(0.25)}$};
    \node [] (unseen2) at (3.5, -3.5) {}
        edge [<-] (sigma2) node at (3, -3.3) {$\predhat{y}_t^{(0.5)}$};
    \node [] (unseen1) at (3.5, -4.5) {}
        edge [<-] (sigma3) node at (3, -4.3) {$\predhat{y}_t^{(0.75)}$};
    \node [] (unseen1) at (3.5, -5.5) {}
        edge [<-] (sigma4) node at (3, -5.3) {$\predhat{y}_t^{(0.95)}$};
    
    \end{tikzpicture}
    }
    \caption{$DNN^{Mul}_1$} \label{fig:dnn_mv1}
    \end{subfigure}
    \begin{subfigure}{0.45\linewidth}
    \scalebox{0.8}{
    \begin{tikzpicture}[x=1.5cm, y=1.5cm, >=stealth]
    
    \node [align=center, above] at (0, -1) {}; 
    \node [align=center, above] at (2, -1) {}; 
    \node [align=center, above] at (4, -1) {}; 
    
    \node [neuron] (input0) at (0, -2) {$x_{t,1}$};
    \node [neuron] (input1) at (0, -3) {$x_{t,2}$};
    \node [missing]  (missing1) at (0, -4) {};
    \node [neuron] (inputN) at (0, -5) {$x_{t,n}$};
    
    \node [neuron] (sigma0) at (2, -1.5) {$\Sigma$}
        edge [<-] (input0) 
        edge [<-] (input1) 
        edge [<-] (inputN);
    \node [neuron] (sigma1) at (2, -2.5) {$\Sigma$}
        edge [<-] (input0) 
        edge [<-] (input1) 
        edge [<-] (inputN);
    \node [neuron] (sigma2) at (2, -3.5) {$\Sigma$}
        edge [<-] (input0) 
        edge [<-] (input1) 
        edge [<-] (inputN);
    \node [neuron] (sigma3) at (2, -4.5) {$\Sigma$}
        edge [<-] (input0) 
        edge [<-] (input1) 
        edge [<-] (inputN);
    \node [neuron] (sigma4) at (2, -5.5) {$\Sigma$}
        edge [<-] (input0) 
        edge [<-] (input1) 
        edge [<-] (inputN);
    
    \node [neuron] (sigma20) at (4, -1.5) {$\Sigma$}
        edge [<-] (sigma0) 
        edge [<-] (sigma1) 
        edge [<-] (sigma2)
        edge [<-] (sigma3)
        edge [<-] (sigma4);
    \node [neuron] (sigma21) at (4, -2.5) {$\Sigma$}
        edge [<-] (sigma0) 
        edge [<-] (sigma1) 
        edge [<-] (sigma2)
        edge [<-] (sigma3)
        edge [<-] (sigma4);
    \node [neuron] (sigma22) at (4, -3.5) {$\Sigma$}
        edge [<-] (sigma0) 
        edge [<-] (sigma1) 
        edge [<-] (sigma2)
        edge [<-] (sigma3)
        edge [<-] (sigma4);
    \node [neuron] (sigma23) at (4, -4.5) {$\Sigma$}
        edge [<-] (sigma0) 
        edge [<-] (sigma1) 
        edge [<-] (sigma2)
        edge [<-] (sigma3)
        edge [<-] (sigma4);
    \node [neuron] (sigma24) at (4, -5.5) {$\Sigma$}
        edge [<-] (sigma0) 
        edge [<-] (sigma1) 
        edge [<-] (sigma2)
        edge [<-] (sigma3)
        edge [<-] (sigma4);
    
    \node [] (unseen0) at (5.5, -1.5) {}
        edge [<-] (sigma20) node at (5, -1.3) {$\predhat{y}_t^{(0.05)}$};
    \node [] (unseen1) at (5.5, -2.5) {}
        edge [<-] (sigma21) node at (5, -2.3) {$\predhat{y}_t^{(0.25)}$};
    \node [] (unseen2) at (5.5, -3.5) {}
        edge [<-] (sigma22) node at (5, -3.3) {$\predhat{y}_t^{(0.5)}$};
    \node [] (unseen1) at (5.5, -4.5) {}
        edge [<-] (sigma23) node at (5, -4.3) {$\predhat{y}_t^{(0.75)}$};
    \node [] (unseen1) at (5.5, -5.5) {}
        edge [<-] (sigma24) node at (5, -5.3) {$\predhat{y}_t^{(0.95)}$};
    
    \end{tikzpicture}    
    }
    \caption{$DNN^{Mul}_2$} \label{fig:dnn_mv2}
    \end{subfigure}
    \caption{Multivariate deep Quantile Regression models for any OD pair. $x_{t,1}, \dots, x_{t,n}$ are input features at lag $t$.}
    \label{fig:deep_qr}
\end{figure}
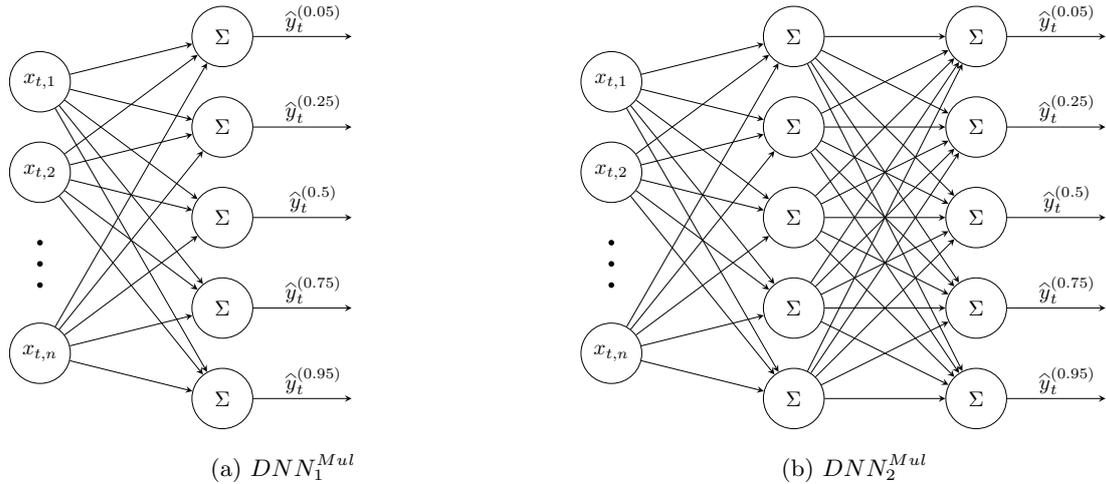

Next, we try multivariate variants of the DNN models which we built earlier.
These variants process multiple quantiles together, and so take less time to train and optimize than the earlier univariate, independent models.

Our first deep multivariate model $DNN_1^{Mul}$ is as illustrated in \fgrref{fig:dnn_mv1}, such that for each OD pair independently, all quantiles are modeled together. 
We again partition $T_{train}$ as $T_{opt}^{train}, T_{opt}^{val}, T_{opt}^{test}$, and use Bayesian Optimization to find the best learning rate in $\left[10^{-8}, 1\right]$, common to all OD pairs. The performance of $DNN_1^{Mul}$ in \tblref{tab:qr_performance} is seen to be worse than $LQR_5^{Ind}$, but better than the univariate $DNN^{Ind}$ models.

Modeling multiple quantiles together can thus result in improved performance. 
We next try to further improve performance by introducing a hidden layer between the input and the output layers, as in \fgrref{fig:dnn_mv2}, so that different quantiles share some trainable parameters. 
\tblref{tab:qr_performance} shows that this model, which we name $DNN_2^{Mul}$, performs badly.

\subsection{Gradient Boosting on Multiple Time Series}

Finally, we fit Gradient Boosting (GBoost) models on the time series of all OD pairs together, independently for each $q \in Q$. As in Section \ref{sec:gboost}, we use Bayesian Optimization to find the best hyper-parameters, and summarize results in \tblref{tab:qr_performance}.

First, we optimize a single set of hyper-parameters, common to all $q \in Q$. This results in model $GBoost_1^{Mul}$, which outperforms $LQR_4^{Ind}$, our previously best performing prediction model. That is, compared to $LQR_4^{Ind}$, $GBoost_1^{Mul}$ achieves better total $MTL$ and significantly better mean $\mil$, with a small decrease in mean $\icp$. 
Next, we optimize hyper-parameters independently for each $q \in Q$.
The resulting model $GBoost_2^{Mul}$ achieves similar total $MTL$ and better mean $\mil$ vs. $LQR_4^{Ind}$, but does not improve over $GBoost_1^{Mul}$.
In summation, $GBoost_1^{Mul}$ yields the best prediction quality among all QR models.

\subsection{Conclusions on Demand Prediction with Quantile Regression}

We have experimented with four types of QR models for predicting distributions of future demand: naive Historical Percentiles, Linear QR, Deep QR, and Gradient Boosting QR. 
Our experiments include independent models for each OD pair and each $q \in Q$, as well as multivariate models which simultaneously process multiple OD pairs or multiple quantiles.
\tblref{tab:qr_performance} summarizes the experiment results. 

Historical Percentiles performs rather poorly, which indicates that the data at hand does not follow a simple repeating pattern.
For Linear and Deep QR, independent models mostly yield better prediction quality than multivariate models.
For Gradient Boosting QR, however, multivariate models outperform independent models, such that $GBoost_1^{Mul}$ is the overall best performing prediction model.

In conclusion, model $GBoost_1^{Mul}$ can exploit spatio-temporal dependencies in the data to some extent.
Indeed, \fgrref{fig:corr_od_pairs} shows that for some OD pairs, the time series of movement counts are strongly correlated.

\subsection{Online Demand Prediction}

The models we have developed for demand prediction are data-driven, and can be trained offline on historical data.
Given online data for recent lags, it takes only a few seconds to apply the trained models and obtain predictive distributions for the next lag.
As new data accumulates over time, the models can be re-trained offline, e.g., once every few days.

\section{Supply Optimization} \label{sec:supplyopt}

So far, we have described how our framework yields predictive distributions of marginal demand.
In this Section, we devise a method for transit network design under such demand stochasticity.
As explained in \ref{sec:stocTRNDP}, this moves us from the context of the classic, static TRNDP, where demands take on deterministic values, to stochastic TRNDP, where demands follow some distributions.

Our solution approach to \emph{stochastic} TRNDP is scenario-based: by independently drawing values for the parameters from their distributions, we generate multiple \emph{static} TRNDP instances, which we then solve independently.
Each solution provides an optimal route and service frequency for the corresponding instance.
Finally, we combine all solutions into an overall optimal solution.

Sampling from the \emph{joint} distribution of parameters is made possible by constructing a Gaussian copula, which can join the \emph{marginal} distributions.
In the following Sections, we first explain how to build the copula, then describe our formulation of the optimization problem instances, and finally describe how to combine them into an overall solution, which is optimal in expectation.
These steps are summarized in Algorithm \ref{alg:opt}.

\begin{algorithm}[ht]
\DontPrintSemicolon
\KwIn{Marginal predictive distributions $\hat{y}_{t,OD_1}, \dots, \hat{y}_{t,OD_N}$ (Sec. \ref{sec:qr}) ; copula $H$ (Sec. \ref{sec:gaussian_copula}) ; number of samples $k$.}
\KwOut{Optimal solution $X_t^*$.} Using $H$, draw $k$ independent samples $s_{t,1}, \dots, s_{t,k}$ from the joint distribution of $\hat{y}_{t,OD_1}, \dots, \
\hat{y}_{t,OD_N}$ (Sec. \ref{sec:sample}).\;
\ForEach{$s_{t,i} \textbf{ in } s_{t,1}, \dots, s_{t,k}$}{
    Calculate optimal solution $X_{t,i}^*$ by solving the linear program in Sec. \ref{sec:linprog} for $s_{t,i}$.
}
\Return the most frequent element among $X_{t,1}^*, \dots, X_{t,k}^*$.
\caption{Supply optimization at lag $t$}
\label{alg:opt}
\end{algorithm}

\subsection{Sampling via Gaussian Copula} \label{sec:copula}

To obtain multiple instances of the stochastic TRNDP problem, we need to be able to jointly sample from the marginal distributions of the parameters.
In order to achieve joint sampling of the distributions, we require a function that pieces together the individual distributions to form a joint distribution.
The function, in essence, retains the correlation structure between the different parameters, and provides us with a mean of obtaining joint samples.
We next define the Gaussian copula and describe how our framework uses it for stochastic optimization.

\subsubsection{Gaussian Copula} \label{sec:gaussian_copula}
Let $F_1(x), \dots, F_n(x)$ be the marginal cumulative distribution functions (CDFs) of demand parameters for each OD pair $X_1, \dots, X_n$, respectively. 
The random vector $\left(X_1, \dots, X_n\right)$ is distributed per the joint cumulative distribution function (CDF) of its components:
\begin{equation}
\label{eq:H1}
H(x_1, \dots, x_n) = \operatorname{Pr} (X_1 \leq x_1, \dots, X_n \leq x_n).
\end{equation}
If each of $F_1(x), \cdots, F_n(x)$ is continuous, then by Sklar's theorem, $H$ is uniquely defined as a function of these marginal CDFs, namely
\begin{equation} 
\label{eq:H2}
H(x_1, \dots, x_n) = C\left(F_1(x_1), \dots, F_n(x_n)\right).
\end{equation}
$C$ is then called a \emph{copula}, and by using this formulation, we decouple the correlation structure in $\left(X_1, \dots, X_n\right)$ from the individual marginals. A commonly used type of copula is the Gaussian copula,
\begin{equation} \label{eq:gaussian_copula}
C_G(u_1, \dots, u_n) = \Phi(u_1, \dots, u_n; \Sigma)
\end{equation}
where $\Phi$ is the joint CDF of the zero-mean multivariate normal distribution with covariance matrix $\Sigma$.  

We can use the Gaussian copula to obtain samples from the joint distribution of $(X_1, \cdots, X_n)$. To obtain such a sample, we first draw $s = (s_1, \cdots, s_n)$ from $C_G$ (Eq. \ref{eq:gaussian_copula}). Then, using the multivariate normal CDF $\Phi_{\Sigma}$, 
we transform $s$ into $\phi = \left(\Phi_{\Sigma}(s_1), \cdots, \Phi_{\Sigma}(s_n)\right)$. Finally, we map $\phi$ to the desired sample through the inverses of the marginal CDFs, as
\begin{equation}
\label{eq:C}
\left(F_1^{\inv} \Phi_{\Sigma}(s_1), \cdots, F_n^{\inv} \Phi_{\Sigma}(s_n)\right).
\end{equation}

\subsubsection{Covariance Matrix for the Gaussian Copula} \label{subsec:predopt}

In this study, we assume that only travel demands are stochastic, while other parameters, such as travel time between nodes, remain constant over time.
To construct the copula, we assume for simplicity that the correlation structure of the data is time-invariant. As such, it suffices to compute the covariance matrix of the copula offline once, based on historical data. 
We note, however, that this computation can be efficiently maintained online as new data becomes known, so that the copula retains the updated state of correlation.


To construct the covariance matrix for the Gaussian copula $C_G$ in our framework, we use the observed movement counts in $T_{train}$, as following.
First, for the $n$ demand nodes in the TRNDP graph $G$, we calculate for each OD pair its empirical marginal CDF $F_{od}$, based on all its historically observed movement counts in $T_{train}$.
For each $t \in T_{train}$, we then collect the corresponding observations of all OD pairs into a travel demand vector of size $n^2$,
\begin{equation}
    \mathbf{d}^{(t)} = \left( d_{11}^{(t)}, \cdots, d_{1n}^{(t)}, d_{21}^{(t)}, \cdots, d_{2n}^{(t)}, \cdots, d_{n1}^{(t)}, \cdots, d_{nn}^{(t)} \right)
\end{equation}
where the travel demand from any demand node to itself, i.e. $d_{ii}$ for $i=[1,n]$, is $0$. 
Next, we transform $\mathbf{d}^{(t)}$ element-wise into the Gaussian layer as follows:
\begin{equation}
\label{eq:d}
    \tilde{d}_{od}^{(t)} = 
    \begin{cases}
        \Phi^{\inv} F_{od}\left(d_{od}^{(t)}\right),  & o \neq d\\
        0,                           & o = d
    \end{cases}
\end{equation}
where $\Phi$ is the standard univariate normal CDF.

Finally, we collect the transformed vectors for all $t \in T_{train}$ into a matrix $\mathbf{\tilde{D}}$, and calculate the covariance matrix in the transformed space as:
\begin{equation}
\label{eq:cov}
\operatorname{Cov}(\mathbf{\tilde{D}}) = \mathbb{E}\left[ (\mathbf{\tilde{D}} - \mu_{\mathbf{\tilde{D}}}) (\mathbf{\tilde{D}} - \mu_{\mathbf{\tilde{D}}})^T \right]
\end{equation}
where $\mu_{\mathbf{\tilde{D}}}$ is given by the column-wise expectation of the matrix $\mathbf{\tilde{D}}$.
$\operatorname{Cov}(\mathbf{\tilde{D}})$ is then the covariance matrix of the Gaussian copula $C_G$

\subsection{TRNDP as a minimum cost flow problem} \label{sec:maxutilitymaxflowformulation}

To solve TRNDP and the frequency determination problem simultaneously, we now rigorously formulate it as a minimum cost flow problem.

\begin{figure}[ht]
    \centering
    \scalebox{0.8}{
    \begin{tikzpicture}

    \node [neuron] (od1) at (0,-3) {$OD_1$};
    \node [neuron] (od2) at (0,-6.5) {$OD_2$};
    \node [neuron] (od3) at (0,-11) {$OD_{|\mathcal{N}|}$};
    
    \node [neuron] (c0) at (4,0.5) {$R_0$}
        edge [<-] (od1)
        edge [<-] (od2)
        edge [<-] (od3);
    \node [neuron] (c1) at (4,-3) {$R_1$}
        edge [<-] (od1)
        edge [<-] (od2)
        edge [<-] (od3);
    \node [neuron] (c2) at (4,-6.5) {$R_2$}
        edge [<-] (od1)
        edge [<-] (od2)
        edge [<-] (od3);
    \node [neuron] (c3) at (4,-11) {$R_C$}
        edge [<-] (od1)
        edge [<-] (od2)
        edge [<-] (od3);
    
    \node [neuron] (c1n1) at (8,-2) {$N_{11}$}
        edge [<-] (c1);
    \node [neuron] (c1n3) at (8,-4) {$N_{1K}$}
        edge [<-] (c1);
    \node [neuron] (c2n1) at (8,-5.5) {$N_{21}$}
        edge [<-] (c2);
    \node [neuron] (c2n3) at (8,-7.5) {$N_{2K}$}
        edge [<-] (c2);
    \node [neuron] (c3n1) at (8,-10) {$N_{C1}$}
        edge [<-] (c3);
    \node [neuron] (c3n3) at (8,-12) {$N_{CK}$}
        edge [<-] (c3);
        
    \node [neuron] (dummy) at (12, -7) {D}
        edge [<-] (c1n1)
        edge [<-] (c1n3)
        edge [<-] (c2n1)
        edge [<-] (c2n3)
        edge [<-] (c3n1)
        edge [<-] (c3n3);
    
    \node [missing] (m1) at (0,-8.75) {};
    \node [missing] (m1) at (4,-8.75) {};
    \node [missing] (m1) at (8,-8.75) {};
    \node [missingsm] (m1) at (8,-3) {};
    \node [missingsm] (m1) at (8,-6.5) {};
    \node [missingsm] (m1) at (8,-11) {};
    
    \draw[thick,dotted] (2.25,2.5) rectangle (5.75,-13);
    \draw[thick,dotted] (6.25,2.5) rectangle (9.75,-13);
    
    \node[text width=1.5cm] at (4,2) {\textbf{Stage 1}};
    \node[text width=1.5cm] at (8,2) {\textbf{Stage 2}};
    
    \draw[->] (c0.east) -| (dummy.north);
    
    \end{tikzpicture}
    }
    \caption{Minimum cost flow formulation for transit route network design problem.} 
    \label{fig:maxflowformulation}
\end{figure}
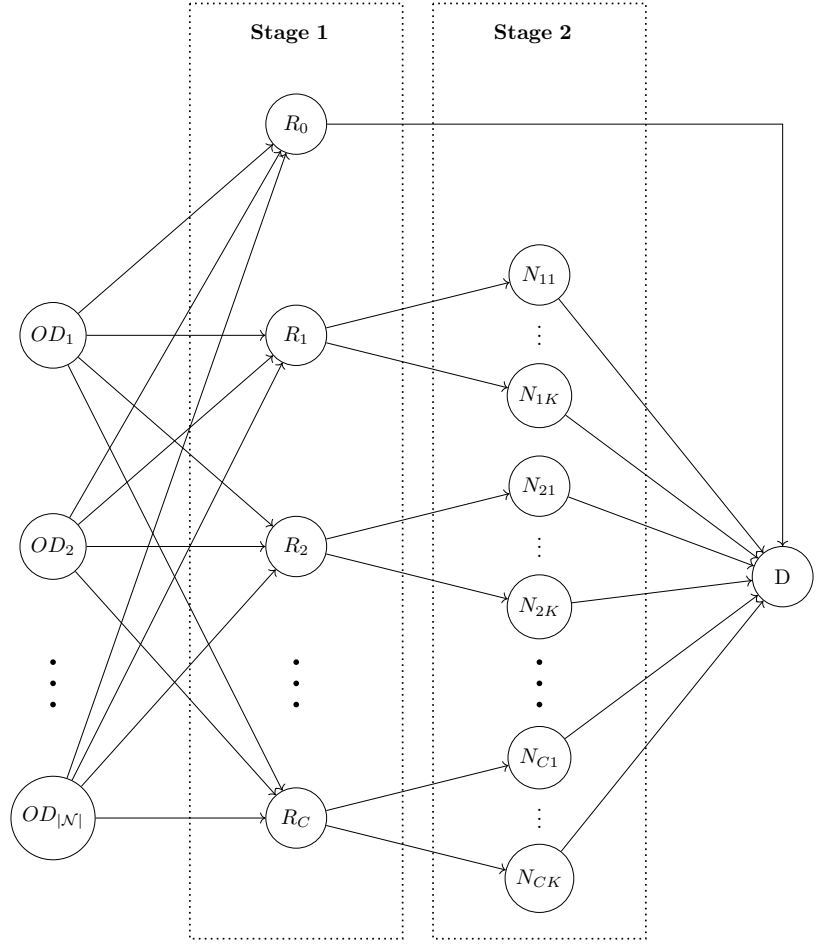

In \fgrref{fig:maxflowformulation}, each OD pair is represented by a node in $\left\{OD_1, \cdots, OD_{|\mathcal{N}|} \right\}$.
They are followed by nodes $\left\{R_0, R_1, \cdots, R_C\right\}$ which represent the candidate routes ; $R_0$ represents the "walking route", which passengers can take if bus capacity is exceeded, or if walking is quicker than taking the bus.
Finally, we have nodes $\left\{ N_{11}, \cdots, N_{1K}, \cdots, N_{C1}, \cdots, N_{CK} \right\}$, per the number of buses assigned to the candidate routes.

In the minimum cost flow formulation, the flow on the edges represent demand flows.
In stage one, when demands flow from an OD node to a route node, the passengers travelling between that OD pair are assigned to the route.
In stage two, the total demands flowing to a route node, flows into a single node that indicates the number of buses allocated to that route. 
For instance, if the demands from $R_1$ flow into node $N_{11}$, it means that the route $R_1$ is allocated one bus.

As in all minimum cost flow problem, we have source nodes and sink nodes where the flows are respectively generated and terminated.
In this case, the source nodes are the OD nodes $\left\{ OD_1, \cdots, OD_{|\mathcal{N}|} \right\}$.
The only sink node is a dummy node $D$, which is added to the network graph to force flows through the network.

The edges are weighted by the time savings achieved for each passenger that travels on a particular route, whereas the objective is to maximize the flow of demands from the source nodes to a dummy sink node node $D$.
Thus, this formulation is equivalent to solving the TNDFS to give the maximum total travel time savings.

\subsubsection{Assumptions}

We employ the following assumptions in the minimum cost flow formulation.
\begin{itemize}
    \item \textit{Passenger demands}: Exogenous demand for each OD pair is either given by the ground truth demands for each hour of the day, or by sampling from the predicted demand distribution.
    \item \textit{Walking time}: The walking time between any two nodes is given by the Manhattan distance between the two nodes and a fixed walking speed.
    \item \textit{Vehicle travel time}:
    For all routes, the travel time between any two consecutive bus stops is exogenously given and fixed. 
    It is assumed that the dwell time is insignificant and thus not considered in this work. Nevertheless, a constant dwell time can easily be incorporated into the formulations.
    \item \textit{Choice behaviour}:
    It is assumed that passengers choose one of several routes to maximize their expected travel time savings compared with respect to walking time.
    \item \textit{Candidate route set}:
    For simplicity, the candidate route set is made up of all possible permutations of the bus stop nodes. While a multitude of candidate route set generation procedures exist in the literature (e.g., \cite{CIPRIANI20123, KILIC201421}), it is not the main focus of this work. 
    Instead, we focus on showing that given demand predictions and a good supply optimization method, it is possible to combine them with a copula to achieve robust optimization.
\end{itemize}

\subsubsection{Linear Program Formulation} \label{sec:linprog}

Let us now formulate the maximum flow problem (\fgrref{fig:maxflowformulation})  as a linear program, which we can then solve through sampling from the predictive distributions.
The linear program uses notations as detailed in \tblref{tab:notations}.

\renewcommand{\arraystretch}{0.85}

\begin{table}[htb!]
    \centering
    \begin{tabularx}{\textwidth}{Q{2cm} >{\setlength{\baselineskip}{0.8\baselineskip}}Z}
        \toprule
        Notation & Definition\\
        \midrule
        \multicolumn{2}{l}{\textbf{Sets andIndices}}\\
        $\mathcal{N}$         & Set of origin and destination nodes.\\
        $o$                   & Origin node.\\
        $d$                   & Destination node.\\
        $c$                   & Candidate route.\\
        \multicolumn{2}{l}{\textbf{Parameters}}\\
        $\beta_{odc}^{(1)}$ & Utility for the transit assignment stage, given by the time savings from taking a ride on candidate route $c$ compared to walking directly from origin $o$ to destination $d$, for each passenger traveling from origin $o$ to destination $d$.\\
        $\beta_{ck}^{(2)}$  & Utility for the buses allocation stage,  given by the negative of the average waiting time for candidate route $c$, with $k$ buses allocated to the route.\\
        $\tau_c$            & Route length of candidate route $c$.\\
        $\lambda_{od}$      & Total travel demand from origin $o$ to destination $d$.\\
        $W_{od}$       & Walk time between origin $o$ and destination $d$.\\
        $B_{odc}$      & Ride time on candidate route $c$ for passengers traveling from origin $o$ to destination $d$.\\
        $W_{odc}^{\prime}$ & Total walk time from origin $o$ to boarding node on candidate route $c$, and walk time from alighting node on candidate route $c$ to destination $d$.\\
        $f_{ck}$            & Frequency of candidate route $c$ when allocated $k$ buses.\\
        $\gamma$            & Capacity of each bus.\\
        $\nu$               & Number of routes permitted.\\
        $C$                 & Number of candidate routes.\\
        $K$                 & Fleet size.\\
        \multicolumn{2}{l}{\textbf{Decision variables}}\\
        $X_{odc}^{(1)}$     & Demands flowing from node $OD_{od}$ to node $R_c$.\\
        $X_{ck}^{(2)}$      & Demands flowing from node $R_c$ to node $N_{ck}$\\
        $\tilde{X}_{ck}$    & Equals to 1 if $k$ buses are allocated to candidate route $c$.\\
        \bottomrule
    \end{tabularx}
    \caption{Notations used in the minimum cost flow formulation.}
    \label{tab:notations}
\end{table}

In the linear program formulation, the utility to be maximized is total time savings of all passengers, namely:
\begin{equation}
\max \q \sum\limits_{\aoD} \sum\limits_{\adD} \sum\limits_{\ac} \sum\limits_{\ak}\beta_{odc}^{(1)}X_{odc}^{(1)} + \beta_{ck}^{(2)}X_{ck}^{(2)}
\end{equation}
where $\beta_{odc}^{(1)}$ and $\beta_{ck}^{(2)}$ represent the edge costs for stages one and two respectively.

In stage one, the edge costs are given by the time savings per passenger, namely:
\begin{equation}
    \beta_{odc}^{(1)} = W_{od} - B_{odc} - W_{odc}^{\prime} \, , \, \forall \aodD \q \ac
\end{equation}
Note that $\beta_{odc}^{(1)}$ does not take into account for waiting time at bus stops.
Instead, wait times are represented through $\beta_{ck}^{(2)}$, the average waiting time for candidate route $c$ when allocated $k$ buses, defined as:
\begin{equation}
    \beta_{ck}^{(2)} = \frac{\tau_{c}}{k} \, , \, \forall \aodD
\end{equation}
Next, the edges between all OD nodes and node $R_0$ are assigned zero cost (i.e. zero time savings):
\begin{equation}
    \beta_{od0} = 0.
\end{equation}

The maximization is subject to the following constraints. 
\begin{flalign}
        	& \sum\limits_{\aoD} \sum\limits_{\substack{\adD\\o \neq d}} X_{odc}^{(1)} = \sum\limits_{\ak} X_{ck}^{(2)}
\label{A1}	& \forall \ac \qq \\
        	& \sum\limits_{0 \leq c \leq C} X_{odc}^{(1)} = \lambda_{od}
\label{A2}	& \forall \aodD \q o \neq d \qq \\
        	& \tilde{X}_{ck} = \begin{cases}1, &X_{ck}^{(2)} > 0\\0, &X_{ck}^{(2)} = 0\end{cases}
\label{A3}	& \forall \ac \q \ak \qq \\
        	& \sum\limits_{\ak} k\tilde{X}_{ck} \leq K
\label{A4}	& \forall \ac \qq \\
        	& \sum\limits_{\ac} \sum\limits_{\ak} k \tilde{X}_{ck} \leq K
\label{A5}	& \\
        	& X_{ck}^{(2)} \leq f_{ck} \gamma
\label{A6}	& \forall \ac \q \ak \qq \\
        	& \sum\limits_{\ac} \sum\limits_{\ak} \tilde{X}_{ck} = \nu
\label{A7}	& 
\end{flalign}

Constraint \eqref{A1} is the flow conservation constraint on the nodes in the route selection stage. The total flow out of any given OD node is made equal to the demands on the OD pair with constraint \eqref{A2}. In constraint \eqref{A3}, the binary variable $\tilde{X}_{ck}$ is defined to determine if the link connecting nodes $R_c$ and $N_{ck}$ is allocated any flow. Constraint \eqref{A4} ensures that the number of buses allocated to each candidate route does not exceed the fleet size, while constraint \eqref{A5} sets the maximum total number of buses allocated to all routes. Constraint \eqref{A6} is the capacity constraint, while \eqref{A7} sets the maxmimum number of routes allowed.

\subsubsection{Solution through Sampling} \label{sec:sample}

We now describe the solution method for the linear program, for any given time $t$ and prediction model $\cal{M}$.
Indices $t$ and $\cal{M}$ are dropped in the following description for conciseness.

Given predicted demands distributions in the form of quantiles $\left(\hat{y}_{od}^{(q)}\right)_{q\in{Q}}$, we have a random variable for the travel demand of each OD pair which follows the empirical distributions defined by the predicted quantiles as follows:
\begin{equation}
    Y_{od} \sim \mathcal{E}\left(\hat{y}_{od}^{(q)}\right) \q,\q q \in Q    
\end{equation}
where $\mathcal{E}$ represents the empirical distribution with Cumulative Distribution Function $F_{\mathcal{E}}$, defined as:
\begin{equation}
    F_{\mathcal{E}}(y) =
    \begin{cases}
    0 & ,y < 0\\
    \frac{y-\hat{y}^{(q_i)}}{\hat{y}^{(q_{i+1})} - \hat{y}^{(q_i)}} + q_i & ,y \in \left(\hat{y}^{(q_i)}, \hat{y}^{(q_{i+1})}\right] \\
    1 & ,y > \hat{y}^{(q_{|Q|})}
    \end{cases}
\end{equation}

By definition, $\hat{y}^{(0)}$  and $\hat{y}^{(1)}$ are given the values of $0$ and $\hat{y}^{(|Q|)} + \hat{y}^{(q_1)}$ respectively.
Then by using the Gaussian copula $H = \Phi_{\Sigma}$ constructed in Section \ref{sec:copula}, we sample $k$ $M$-dimensional sample demand vectors $\left\{\vec{\lambda}_i | 1 \leq i \leq k\right\}$ as such:
\begin{equation}
    \vec{\lambda}_i = \begin{bmatrix}Y_1^{-1}\Phi_{\Sigma}(s_{i,1}) & \cdots & Y_M^{-1}\Phi_{\Sigma}(s_{i,M})\end{bmatrix}
\end{equation}
where $\vec{s}_i = \begin{bmatrix} s_{i,1} & \cdots & s_{i,M} \end{bmatrix} \leftarrow \Phi_{\Sigma}$ is a vector drawn from the multivariate Gaussian distribution with covariance matrix $\Sigma$.

For each sample demand vector $\vec{\lambda}_i$, we instantiate a corresponding linear program $P_i$, such that $\lambda_{od}$ in constraint (\ref{A2}) is given by the sampled demand vector $\vec{\lambda}_i$. 
We then solve $P_i$ to obtain an optimal solution $X^\ast_i$ for sample $s_i$.
Finally, we aggregate these solutions to obtain the overall optimal solution $X^\ast$, as:
\begin{equation}
    X^\ast = \mode \left[X^\ast_1, \dots, X^\ast_k\right]
\end{equation}

\section{Case Study for Stochastic Optimization} \label{sec:case_sup}

\begin{figure}
\begin{floatrow}
\ffigbox[0.39\textwidth]{
  \includegraphics[width=0.9\linewidth]{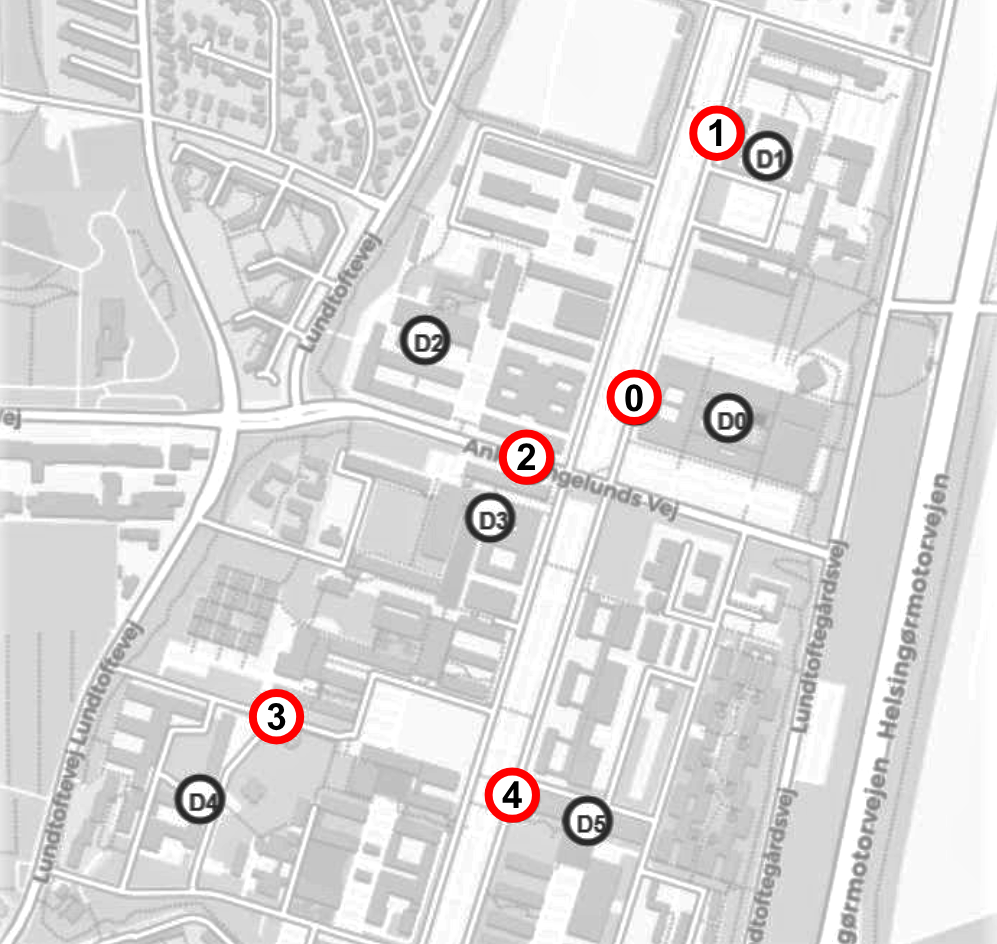}
}{
\caption{Nodes in our case study, labeled with indices. Prefix 'D' indicates a demand node, while other nodes are bus stops.} \label{fig:nodes}
}
\capbtabbox[0.55\textwidth]{
    \begin{tabular}{ll}
        \toprule
        Route Set & Itineraries\\
        \midrule
        $RS1$ & 0-2-0\\
        $RS2$ & 0-2-0, 0-4-0\\
        $RS3$ & 0-2-0, 1-2-1\\
        \bottomrule
    \end{tabular}
}{
  \caption{Commonly recurring solutions in the case study results. Routes in itineraries are given as lists of bus stops in order of bus visitation.} \label{tbl:rs}
}
\end{floatrow}
\end{figure}

So far, we have defined our general supply optimization method.
Let us now apply this method to the case study of autonomous shuttles in the Danish university campus, for which we have developed several demand prediction models in Section \ref{sec:qr}.
Each of these models uses one of various types of Quantile Regression, and yields a marginal predictive distribution for every OD pair in the case study.
Now, we pick several of the best performing models, as per \tblref{tab:qr_performance}, and evaluate each of them by feeding its marginals into the supply optimization method.

In the following Sections, we perform this evaluation for 8-Jan-2018, the first day in the test set, hours 08:00, 09:00, $\dots$, 18:00.
The nodes we use for optimization are described in \fgrref{fig:nodes}, and we set the number of samples at $k = 100$.
First, we compare optimization with ground truth observations vs. optimization with predictive distributions.
Then, we study the robustness of our method against conventional optimization methods.

The optimal itineraries in this case study are sets of routes, where each route is a sequence of bus stops serviced by a single bus.
For brevity, we use the notation in \tblref{tbl:rs} to refer to commonly recurring solutions in the results.
Note, however, that the candidate route set consists not only of the few routes in these solutions, but rather of all possible routes through the $5$ bus stops.

\subsection{Optimization with Ground Truth vs. Predictions}

\renewcommand{\arraystretch}{0.78}

\begin{table*}[ht!]
    \small\addtolength{\tabcolsep}{10pt}
    \resizebox{\linewidth}{!}{
    \begin{tabular}{c l l l l l l l}
    \toprule
    \multirow{2}{*}{Hour} & \multirow{2}{*}{GT} & \multicolumn{6}{c}{Prediction models}\\
    \cmidrule{3-8}
    & & $LQR_4^{Ind}$  & $HP^{Ind}$          & $DNN^{Ind}_1$        & $GBoost_3^{Ind}$     & $DNN_1^{Mul}$        & $GBoost_1^{Mul}$\\
    \midrule
8  & RS1        & RS2 (65) & RS2 (87) & RS2 (67) & RS2 (70) & RS2 (74) & RS2 (100) \\
   &              & * (27)            & * (12)            & RS3 (28) & * (13)            & RS3 (23) &                    \\
   &              &                   &                   & * (5)             &                   & * (3)             &                    \\ \midrule
9  & RS1        & * (98)            & * (100)           & * (100)           & * (100)           & * (99)            & * (99)             \\ \midrule
10 & RS2 & RS1 (100)       & RS1 (100)       & RS1 (100)       & RS1 (99)        & RS1 (100)       & RS1 (100)        \\
   &              &                   &                   &                   & * (1)             &                   &                    \\ \midrule
11 & RS1        & * (92)            & * (97)            & * (93)            & * (98)            & * (95)            & * (65)             \\ \midrule
12 & RS1        & * (91)            & * (93)            & * (94)            & * (100)           & * (97)            & * (92)             \\ \midrule
13 & RS1        & * (100)           & * (100)           & * (100)           & * (99)            & * (100)           & * (100)            \\ \midrule
14 & RS1        & * (98)            & * (98)            & * (94)            & * (65)            & * (94)            & * (97)             \\ \midrule
15 & RS1        & * (96)            & * (99)            & * (97)            & * (99)            & * (98)            & * (99)             \\ \midrule
16 & RS1        & * (90)            & * (92)            & * (85)            & * (94)            & * (88)            & * (96)             \\ \midrule
17 & RS2 & RS1 (97)        & RS1 (100)       & RS1 (93)        & RS1 (97)        & RS1 (91)        & RS1 (100)        \\
   &              & * (3)             &                   & * (7)             & * (3)             & * (9)             &                    \\ \midrule
18 & RS2 & * (97)            & * (100)           & * (97)            & * (98)            & * (95)            & * (92)             \\ \bottomrule
\end{tabular}}
    \caption{Hourly optimal solutions. 
    Column GT lists true optimal solutions, as obtained from ground truth data.
    Other columns list optimal solutions as obtained through demand sampling for several best performing prediction models. 
    Solutions are ranked according to their number of occurrences (in parentheses), and each list terminates at the same route as GT, denoted by an asterisk.}
    \label{tab:opt_gt_vs_pred}
\end{table*}

\tblref{tab:opt_gt_vs_pred} shows that for all prediction models and all hours except 8:00, optimization with predictive distributions yields the \emph{GT solution}, i.e., the same solution as optimization with ground truth observations.
Furthermore, these solutions are obtained with high confidence, i.e., the most frequent solution corresponds to more than $90\%$ of the samples from the predictive distribution.
Hour 08:00, however, is more challenging: the confidence of optimization varies across models, and model $GBoost_1^{Mul}$ completely misses the GT solution.

\begin{figure}[ht!]
    \centering
    \begin{subfigure}[b]{0.49\textwidth}
        \includegraphics[trim={0.9cm 1cm 1cm 1cm}, clip, width=\textwidth]{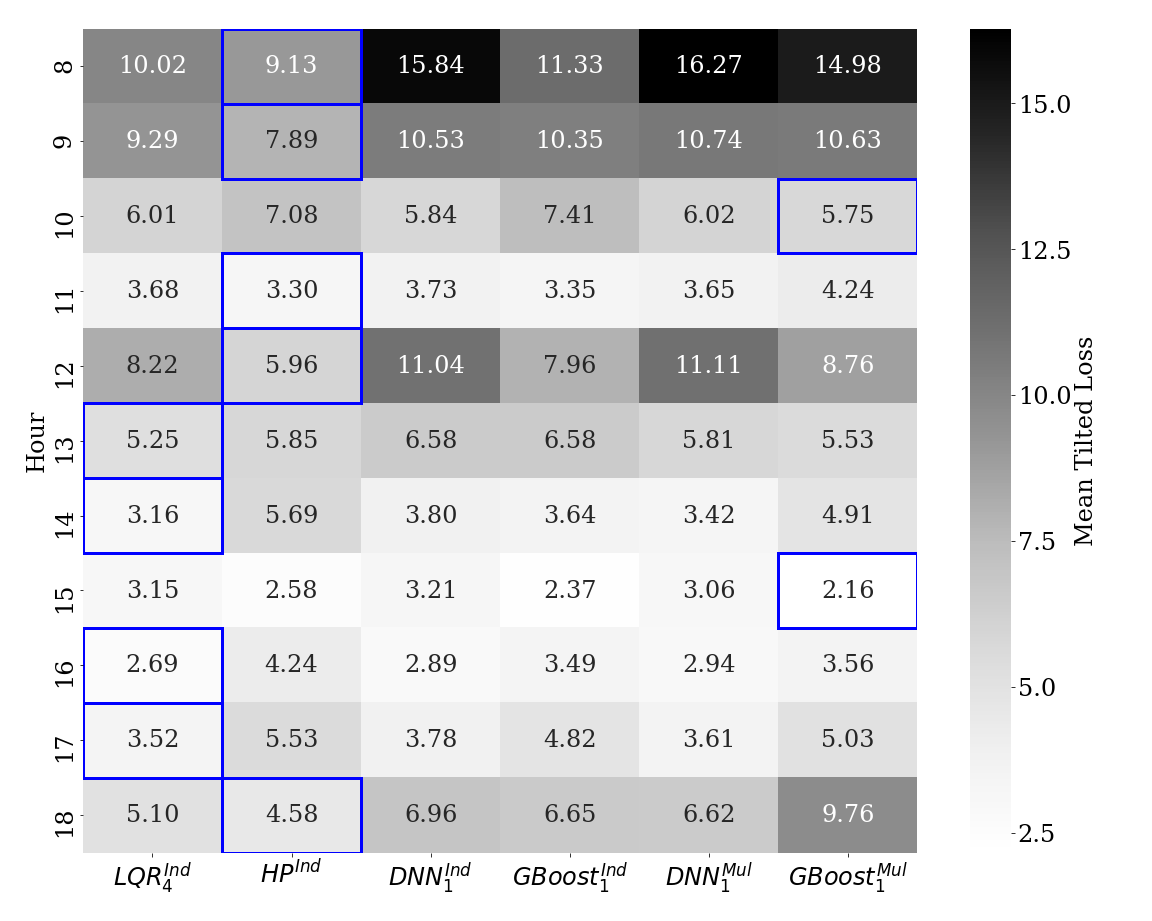}
    \end{subfigure}
    \hfill
    \begin{subfigure}[b]{0.49\textwidth}
        \includegraphics[trim={0.9cm 1cm 1cm 1cm}, clip, width=\textwidth]{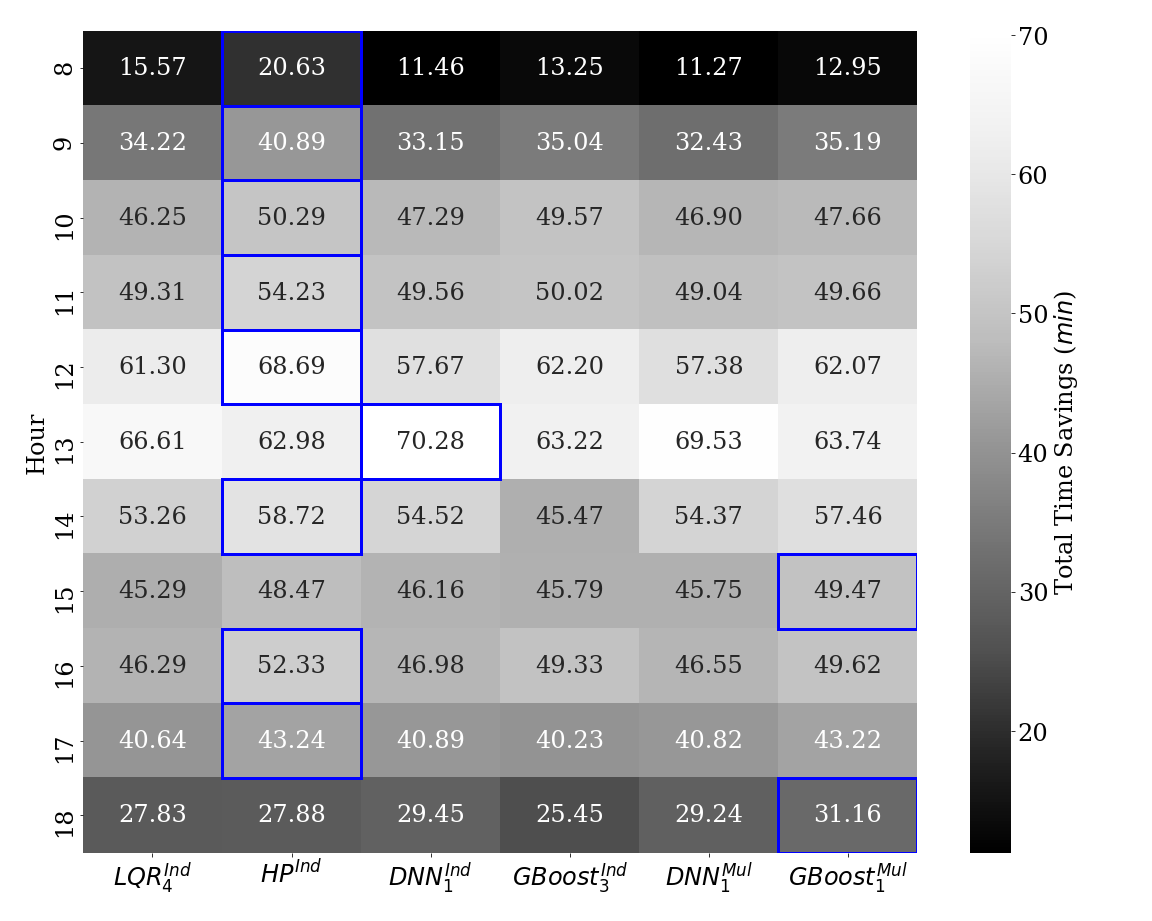}
    \end{subfigure}
    \caption{Prediction vs. optimization quality for best performing prediction models in 2018-Jan-08 08:00$,\dots,$18:00. For each hour and model, the left plot shows Mean Tilted Loss over all OD pairs and $q \in Q$, while the right plot shows average time savings ($min$) over $100$ samples from the corresponding joint predictive distribution. The best value in each hour is framed.}
    \label{fig:pred_vs_opt}
\end{figure}

Earlier, \tblref{tab:qr_performance} showed that the \emph{overall} predictive performance of model $HP^{Ind}$ was considerably worse than $GBoost_1^{Mul}$.
However, in \tblref{tab:opt_gt_vs_pred}, $HP^{Ind}$ often yields better \emph{hourly} optimization performance than $GBoost_1^{Mul}$.
This apparent discrepancy is explained by \fgrref{fig:pred_vs_opt}, which displays the \emph{hourly} predictive performance of each model, measured as hourly Mean Tilted Loss over all OD pairs and $q \in Q$.
We see that for this portion of the test set, $HP^{Ind}$ is indeed the model that most often yields the lowest hourly $MTL$.

Moreover, \fgrref{fig:pred_vs_opt} juxtaposes hourly prediction quality against optimization quality, measured in mean time savings over the $100$ samples from the predictive distribution.
We see that in each hour, the mean time savings of different models are quite close, and $HP^{Ind}$ yields the highest time savings. 
In hour 15:00, $GBoost_1^{Mul}$ is best for both prediction and optimization.
In hours 13:00, $\dots$, 18:00, $LQR_4^{Ind}$ is often the best model for prediction, but not so for optimization.

\subsection{Comparison to Conventional Optimization Methods}

\begin{table*}[ht]
    \resizebox{\linewidth}{!}{
    \begin{tabular}{ccaccaccaccaccaccacc}
        \toprule
        \multirow{2}{*}{Hour} & \multirow{2}{*}{GT} & \multicolumn{3}{c}{$LQR_4^{Ind}$} & \multicolumn{3}{c}{$HP^{Ind}$} & \multicolumn{3}{c}{$DNN_1^{Ind}$} & \multicolumn{3}{c}{$GBoost_3^{Ind}$} & \multicolumn{3}{c}{$DNN_1^{Mul}$} & \multicolumn{3}{c}{$GBoost_1^{Mul}$}\\
        \cmidrule(lr){3-5} \cmidrule(lr){6-8} \cmidrule(lr){9-11} \cmidrule(lr){12-14} \cmidrule(lr){15-17} \cmidrule(lr){18-20}
        & & P & M & R & P & M & R & P & M & R & P & M & R & P & M & R & P & M & R \\
        \midrule
8  & RS1        & RS2 & RS2 & RS2 & RS2 & RS2 & RS2 & RS2 & RS2 & RS2 & RS2 & RS2 & RS2 & RS2 & RS2 & RS2 & RS2 & RS2 & RS2 \\
9  & RS1        & *            & *            & *            & *            & *            & *            & *            & *            & *            & *            & *            & *            & *            & *            & *            & *            & *            & *            \\
10 & RS2 & RS1        & RS1        & RS1        & RS1        & RS1        & RS1        & RS1        & RS1        & RS1        & RS1        & RS1        & RS1        & RS1        & RS1        & RS1        & RS1        & RS1        & RS1        \\
11 & RS1        & *            & *            & *            & *            & *            & *            & *            & RS2 & RS2 & *            & *            & *            & *            & *            & *            & *            & *            & RS2 \\
12 & RS1        & *            & *            & *            & *            & *            & *            & *            & *            & RS2 & *            & *            & *            & *            & *            & *            & *            & *            & *            \\
13 & RS1        & *            & *            & *            & *            & *            & *            & *            & *            & *            & *            & *            & *            & *            & *            & *            & *            & *            & *            \\
14 & RS1        & *            & *            & *            & *            & *            & *            & *            & *            & *            & *            & *            & *            & *            & *            & *            & *            & *            & *            \\
15 & RS1        & *            & *            & *            & *            & *            & *            & *            & *            & *            & *            & *            & *            & *            & *            & *            & *            & *            & *            \\
16 & RS1        & *            & *            & *            & *            & *            & *            & *            & *            & *            & *            & *            & *            & *            & *            & *            & *            & *            & *            \\
17 & RS2 & RS1        & RS1        & RS1        & RS1        & RS1        & RS1        & RS1        & RS1        & RS1        & RS1        & RS1        & RS1        & RS1        & RS1        & RS1        & RS1        & RS1        & RS1        \\
18 & RS2 & *            & *            & *            & *            & *            & *            & *            & *            & *            & *            & *            & *            & *            & *            & *            & *            & *            & *            \\
        \bottomrule
    \end{tabular}}
    \caption{Hourly optimal solutions using different optimization strategies. GT = optimization with ground truth observations. P = proposed predictive optimization framework, M = conventional optimization using median point estimates, R = robust optimization (worst-case optimization). Asterisks denote same solution as GT.}
    \label{tbl:opt_strats}
\end{table*}

Now, we compare our optimization framework with two conventional optimization strategies: 1) using median estimates for the parameters, and 2) worst-case optimization (i.e., robust optimization).
As discussed in Section \ref{sec:stocTRNDP}, both of these methods are commonly used to reduce the full distributions of the parameters into point estimates so that the problems can readily be solved.
For each hour $t$ and prediction model $\cal{M}$, we thus collapse the predictive distribution to point estimates: 1) the $50\%$ quantile for median estimate, and 2) the $95\%$ quantile for worst-case estimate.

The optimization results for all methods are presented in Table \ref{tbl:opt_strats}.
We see that our optimization method performs mostly on par with the conventional optimization strategies in terms of GT solutions.
Furthermore, our framework outperforms the other methods for models $DNN_1^{Ind}$ and $GBoost_1^{Mul}$.
We attribute this gain in performance to the ability of our framework to take in full density estimates for the parameters, as opposed to point estimates in the conventional methods.

\subsection{Online Supply Optimization}

For this case study, it takes only a few seconds to both draw a sample from the joint distribution and solve the corresponding linear program instance.
Because our optimization method does so independently for multiple samples, it is straightforward to parallelize this step online.
Finding the most frequent solution thereafter takes a few seconds as well even among tens of thousands of solutions. 
For case studies of larger magnitude, the NP-hard optimization problem at hand can be approximated by reducing the candidate route set through various techniques (e.g., \cite{CIPRIANI20123, KILIC201421}).
This results in a tractable problem, which can again be solved in real time.

\section{Conclusion} \label{sec:conclusion}

In this study, we present an online framework for optimally adapting supply to demand.
On the demand side, the framework yields density estimates of future demand in the form of distributions.
On the supply side, the framework uses the demand distributions to compute a solution which is optimal in expectation.
This differs from previous approaches, which focus mostly on one of the two sides, or which have either too little or too much robustness built into the optimization method.

The setting we consider in this study is predictive routing for a fleet of vehicles. 
In this setting, there exists a latent spatio-temporal distribution of demand for using the transit service to commute between Origin-Destination (OD) pairs.
Given online information about crowd movements in the serviced area (e.g., via a network of sensors), our framework estimates this demand by predicting the marginal distribution of movement counts for each OD pair.

A key component of our framework is the construction of a Gaussian copula, through which the updated marginal distributions are combined into a joint spatio-temporal demand distribution. This joint distribution allows for supply optimization on a full predictive demand distribution, rather than just point estimates of future demand.

On the supply side, our framework uses a novel, demand-responsive optimization method to select an optimal route and frequency for the transit service. 
This method uses the aforementioned copula to sample from the joint demand distribution, solve a linear optimization program for each sample, and combine the results into an overall optimal solution.
As such, our framework applies generally to any stochastic linear program, where the predictive distribution of parameters can be estimated.

We evaluate our framework through an actual case study of mobility in a university campus, for which we have aggregated counts of movements between buildings, as collected from WiFi records.
On the demand side, we build and test several prediction models, each of which yields marginal distributions of future demand by estimating multiple CDF quantiles.
On the supply side, we then compute an optimal solution on the output of the best performing prediction models.
The results show that our framework often yields a solution equal to the solution obtained a posteriori, i.e., with ground truth observations.
We also show that our framework performs better than conventional methods for route optimization, which do not utilize full predictive distributions.

\subsection{Future Work}

For future work, we wish to further explore our predictive optimization framework, as follows.

\begin{itemize}
    \item On the demand side, we intend to test additional methods for prediction under uncertainty, including Bayesian Inference, and Deep Neural Networks with prediction intervals.
    
    \item On the supply side, we intend to test our robust optimization method against other instances of the stochastic Transit Routing Network Design Problem.
    
    \item We wish to study the performance of our framework with other forms of copula, e.g., Archimedean copulas.
    
    \item We are interested in studying the scalability of our framework as a whole, by applying it to additional case studies with larger data sets.
    
    \item We intend to derive probabilistic bounds for the accuracy of our stochastic optimization method as a function of the number of samples from the joint demand distribution. Similar probabilistic analysis can be found in \cite{hentenryck2009online} for a different class of online stochastic routing problems.
\end{itemize}

\section*{Acknowledgement}

This research was conducted as part of the LINC project funded by the Urban Innovation Action, and has also received funding from the People Programme (Marie Curie Actions) of the European Union’s Horizon 2020 research and innovation programme under the Marie Sklodowska-Curie Individual Fellowship H2020-MSCA-IF-2016, ID number 745673.

\FloatBarrier

\bibliographystyle{elsarticle-num}
\bibliography{references}

\begin{thebibliography}{10}
\expandafter\ifx\csname url\endcsname\relax
  \def\url#1{\texttt{#1}}\fi
\expandafter\ifx\csname urlprefix\endcsname\relax\def\urlprefix{URL }\fi
\expandafter\ifx\csname href\endcsname\relax
  \def\href#1#2{#2} \def\path#1{#1}\fi

\bibitem{ntu2018autonomous}
Z.~Abdullah, Ntu gets new driverless shuttle bus to ferry students across
  campus,
  \url{https://www.straitstimes.com/singapore/transport/ntu-gets-new-driverless-shuttle-bus-to-ferry-students-across-campus}
  (2018).

\bibitem{linc2018largest}
LINC, The largest test of self-driving shuttles in denmark,
  \url{http://lincproject.dk/en/om/} (2018).

\bibitem{driverless2018}
L.~Alton, Driverless cars will impact healthcare for better and worse,
  \url{https://www.healthworkscollective.com/driverless-cars-will-impact-healthcare-better-worse/}
  (2018).

\bibitem{sevtsuk2009mapping}
A.~Sevtsuk, Mapping the mit campus in real time using wifi, in: Handbook of
  Research on Urban Informatics: The Practice and Promise of the Real-Time
  City, IGI Global, 2009, pp. 326--338.

\bibitem{meneses2012large}
F.~Meneses, A.~Moreira, Large scale movement analysis from wifi based location
  data, in: Indoor Positioning and Indoor Navigation (IPIN), 2012 International
  Conference on, IEEE, 2012, pp. 1--9.

\bibitem{gonzalez2008understanding}
M.~C. Gonzalez, C.~A. Hidalgo, A.-L. Barabasi, Understanding individual human
  mobility patterns, nature 453~(7196) (2008) 779.

\bibitem{guo2012discovering}
D.~Guo, X.~Zhu, H.~Jin, P.~Gao, C.~Andris, Discovering spatial patterns in
  origin-destination mobility data, Transactions in GIS 16~(3) (2012) 411--429.

\bibitem{guo2017novel}
Q.~Guo, H.~A. Karimi, A novel methodology for prediction of spatial-temporal
  activities using latent features, Computers, Environment and Urban Systems 62
  (2017) 74--85.

\bibitem{rodrigues2018beyond}
F.~Rodrigues, F.~C. Pereira, Beyond expectation: Deep joint mean and quantile
  regression for spatio-temporal problems, arXiv preprint arXiv:1808.08798.

\bibitem{de2007uncertainty}
G.~De~Jong, A.~Daly, M.~Pieters, S.~Miller, R.~Plasmeijer, F.~Hofman,
  Uncertainty in traffic forecasts: literature review and new results for the
  netherlands, Transportation 34~(4) (2007) 375--395.

\bibitem{rasouli2012uncertainty}
S.~Rasouli, H.~Timmermans, Uncertainty in travel demand forecasting models:
  literature review and research agenda, Transportation letters 4~(1) (2012)
  55--73.

\bibitem{rodrigues2018heteroscedastic}
F.~Rodrigues, F.~C. Pereira, Heteroscedastic gaussian processes for uncertainty
  modeling in large-scale crowdsourced traffic data, Transportation research
  part C: emerging technologies 95 (2018) 636--651.

\bibitem{tsai2009neural}
T.-H. Tsai, C.-K. Lee, C.-H. Wei, Neural network based temporal feature models
  for short-term railway passenger demand forecasting, Expert Systems with
  Applications 36~(2) (2009) 3728--3736.

\bibitem{yang2013sensitivity}
C.~Yang, A.~Chen, X.~Xu, S.~Wong, Sensitivity-based uncertainty analysis of a
  combined travel demand model, Transportation Research Part B: Methodological
  57 (2013) 225--244.

\bibitem{shao2014estimation}
H.~Shao, W.~H. Lam, A.~Sumalee, A.~Chen, M.~L. Hazelton, Estimation of mean and
  covariance of peak hour origin--destination demands from day-to-day traffic
  counts, Transportation Research Part B: Methodological 68 (2014) 52--75.

\bibitem{xue2015short}
R.~Xue, D.~J. Sun, S.~Chen, Short-term bus passenger demand prediction based on
  time series model and interactive multiple model approach, Discrete Dynamics
  in Nature and Society 2015.

\bibitem{yang2019estimating}
Y.~Yang, Y.~Fan, J.~O. Royset, Estimating probability distributions of travel
  demand on a congested network, Transportation Research Part B: Methodological
  122 (2019) 265--286.

\bibitem{li2019robust}
S.~Li, R.~Liu, L.~Yang, Z.~Gao, Robust dynamic bus controls considering delay
  disturbances and passenger demand uncertainty, Transportation Research Part
  B: Methodological 123 (2019) 88--109.

\bibitem{chen2019quantile}
S.~Chen, Quantile regression for duration models with time-varying regressors,
  Journal of Econometrics 209~(1) (2019) 1--17.

\bibitem{khan2019prediction}
N.~Khan, S.~Shahid, L.~Juneng, K.~Ahmed, T.~Ismail, N.~Nawaz, Prediction of
  heat waves in pakistan using quantile regression forests, Atmospheric
  Research 221 (2019) 1--11.

\bibitem{koenker2005quantile}
R.~Koenker, Quantile regression, volume 38 of econometric society monographs
  (2005).

\bibitem{antunes2017review}
F.~Antunes, A.~O’Sullivan, F.~Rodrigues, F.~Pereira, A review of
  heteroscedasticity treatment with gaussian processes and quantile regression
  meta-models, in: Seeing Cities Through Big Data, Springer, 2017, pp.
  141--160.

\bibitem{yang2018power}
Y.~Yang, S.~Li, W.~Li, M.~Qu, Power load probability density forecasting using
  gaussian process quantile regression, Applied Energy 213 (2018) 499--509.

\bibitem{sangnier2016joint}
M.~Sangnier, O.~Fercoq, F.~d'Alch{\'e} Buc, Joint quantile regression in
  vector-valued rkhss, in: Advances in Neural Information Processing Systems,
  2016, pp. 3693--3701.

\bibitem{peled2019model}
I.~Peled, F.~Rodrigues, F.~C. Pereira, Model-based machine learning for
  transportation, in: Mobility Patterns, Big Data and Transport Analytics,
  Elsevier, 2019, pp. 145--171.

\bibitem{mazloumi2011prediction}
E.~Mazloumi, G.~Rose, G.~Currie, S.~Moridpour, Prediction intervals to account
  for uncertainties in neural network predictions: Methodology and application
  in bus travel time prediction, Engineering Applications of Artificial
  Intelligence 24~(3) (2011) 534--542.

\bibitem{khosravi2011comprehensive}
A.~Khosravi, S.~Nahavandi, D.~Creighton, A.~F. Atiya, Comprehensive review of
  neural network-based prediction intervals and new advances, IEEE Transactions
  on neural networks 22~(9) (2011) 1341--1356.

\bibitem{Ceder1986-lk}
A.~Ceder, N.~H.~M. Wilson, Bus network design, Trans. Res. Part B: Methodol.
  20~(4) (1986) 331--344.

\bibitem{Kepaptsoglou2009-sw}
K.~Kepaptsoglou, M.~Asce, M.~Karlaftis, M.~Asce, Transit route network design
  problem: Review, J. Transp. Eng. 135~(8) (2009) 491--505.

\bibitem{ceder_2016}
A.~Ceder, Public transit planning and operation: modelling, practice and
  behavior, CRC Press, 2016.

\bibitem{GUIHAIRE20081251}
V.~Guihaire, J.-K. Hao,
  \href{http://www.sciencedirect.com/science/article/pii/S0965856408000888}{Transit
  network design and scheduling: A global review}, Transportation Research Part
  A: Policy and Practice 42~(10) (2008) 1251 -- 1273.
\newblock \href {http://dx.doi.org/https://doi.org/10.1016/j.tra.2008.03.011}
  {\path{doi:https://doi.org/10.1016/j.tra.2008.03.011}}.
\newline\urlprefix\url{http://www.sciencedirect.com/science/article/pii/S0965856408000888}

\bibitem{FARAHANI2013281}
R.~Z. Farahani, E.~Miandoabchi, W.~Szeto, H.~Rashidi,
  \href{http://www.sciencedirect.com/science/article/pii/S0377221713000106}{A
  review of urban transportation network design problems}, European Journal of
  Operational Research 229~(2) (2013) 281 -- 302.
\newblock \href {http://dx.doi.org/https://doi.org/10.1016/j.ejor.2013.01.001}
  {\path{doi:https://doi.org/10.1016/j.ejor.2013.01.001}}.
\newline\urlprefix\url{http://www.sciencedirect.com/science/article/pii/S0377221713000106}

\bibitem{Cipriani2012-zm}
E.~Cipriani, S.~Gori, M.~Petrelli, Transit network design: A procedure and an
  application to a large urban area, Transp. Res. Part C: Emerg. Technol.
  20~(1) (2012) 3--14.

\bibitem{Fan2008-lg}
W.~Fan, R.~B. Machemehl, A tabu search based heuristic method for the transit
  route network design problem, in: Computer-aided Systems in Public Transport,
  Springer Berlin Heidelberg, 2008, pp. 387--408.

\bibitem{Fan_Wei2006-mj}
{Fan Wei}, {Machemehl Randy B.}, Using a simulated annealing algorithm to solve
  the transit route network design problem, J. Transp. Eng. 132~(2) (2006)
  122--132.

\bibitem{Szeto2011-qj}
W.~Y. Szeto, Y.~Wu, A simultaneous bus route design and frequency setting
  problem for tin shui wai, hong kong, Eur. J. Oper. Res. 209~(2) (2011)
  141--155.

\bibitem{Szeto2014-pn}
W.~Y. Szeto, Y.~Jiang, Transit route and frequency design: Bi-level modeling
  and hybrid artificial bee colony algorithm approach, Trans. Res. Part B:
  Methodol. 67 (2014) 235--263.

\bibitem{Cipriani2012-yr}
E.~Cipriani, S.~Gori, M.~Petrelli, A bus network design procedure with elastic
  demand for large urban areas, Public Transport 4~(1) (2012) 57--76.

\bibitem{Lee_Young-Jae2005-on}
{Lee Young-Jae}, {Vuchic Vukan R.}, Transit network design with variable
  demand, J. Transp. Eng. 131~(1) (2005) 1--10.

\bibitem{Fan_Wei2006-lc}
{Fan Wei}, {Machemehl Randy B.}, Optimal transit route network design problem
  with variable transit demand: Genetic algorithm approach, J. Transp. Eng.
  132~(1) (2006) 40--51.

\bibitem{CIPRIANI20123}
E.~Cipriani, S.~Gori, M.~Petrelli,
  \href{http://www.sciencedirect.com/science/article/pii/S0968090X10001397}{Transit
  network design: A procedure and an application to a large urban area},
  Transportation Research Part C: Emerging Technologies 20~(1) (2012) 3 -- 14,
  special issue on Optimization in Public Transport+ISTT2011.
\newblock \href {http://dx.doi.org/https://doi.org/10.1016/j.trc.2010.09.003}
  {\path{doi:https://doi.org/10.1016/j.trc.2010.09.003}}.
\newline\urlprefix\url{http://www.sciencedirect.com/science/article/pii/S0968090X10001397}

\bibitem{An2015-tv}
K.~An, H.~K. Lo, Robust transit network design with stochastic demand
  considering development density, Trans. Res. Part B: Methodol. 81 (2015)
  737--754.

\bibitem{Laporte2010-ud}
G.~Laporte, J.~A. Mesa, F.~Perea, A game theoretic framework for the robust
  railway transit network design problem, Trans. Res. Part B: Methodol. 44~(4)
  (2010) 447--459.

\bibitem{Lou2009-et}
Y.~Lou, Y.~Yin, S.~Lawphongpanich, Robust approach to discrete network designs
  with demand uncertainty, Transp. Res. Rec. 2090~(1) (2009) 86--94.

\bibitem{An2016-bx}
K.~An, H.~K. Lo, Two-phase stochastic program for transit network design under
  demand uncertainty, Trans. Res. Part B: Methodol. 84 (2016) 157--181.

\bibitem{Amiripour_S_M_Mahdi2014-om}
{Amiripour S. M. Mahdi}, {Ceder Avishai (Avi)}, {Mohaymany Afshin Shariat},
  Hybrid method for bus network design with high seasonal demand variation, J.
  Transp. Eng. 140~(6) (2014) 04014015.

\bibitem{Stein1978DAR}
D.~M. Stein, \href{http://dx.doi.org/10.1287/trsc.12.3.232}{Scheduling
  dial-a-ride transportation systems}, Transportation Science 12~(3) (1978)
  232--249.
\newblock \href {http://dx.doi.org/10.1287/trsc.12.3.232}
  {\path{doi:10.1287/trsc.12.3.232}}.
\newline\urlprefix\url{http://dx.doi.org/10.1287/trsc.12.3.232}

\bibitem{cortes2009hybrid}
C.~E. Cort{\'e}s, D.~S{\'a}ez, A.~N{\'u}{\~n}ez, D.~Mu{\~n}oz-Carpintero,
  Hybrid adaptive predictive control for a dynamic pickup and delivery problem,
  Transportation Science 43~(1) (2009) 27--42.

\bibitem{powell2003stochastic}
W.~B. Powell, H.~Topaloglu, Stochastic programming in transportation and
  logistics, Handbooks in operations research and management science 10 (2003)
  555--635.

\bibitem{ferrucci2013pro}
F.~Ferrucci, S.~Bock, M.~Gendreau, A pro-active real-time control approach for
  dynamic vehicle routing problems dealing with the delivery of urgent goods,
  European Journal of Operational Research 225~(1) (2013) 130--141.

\bibitem{alonso2017predictive}
A.-M. Javier, W.~Alex, R.~Daniela, Predictive routing for autonomous
  mobility-on-demand systems with ride-sharing, in: Intelligent Robots and
  Systems (IROS), 2017 IEEE/RSJ International Conference on, IEEE, 2017, pp.
  3583--3590.

\bibitem{iglesias2018data}
R.~Iglesias, F.~Rossi, K.~Wang, D.~Hallac, J.~Leskovec, M.~Pavone, Data-driven
  model predictive control of autonomous mobility-on-demand systems, in: 2018
  IEEE International Conference on Robotics and Automation (ICRA), IEEE, 2018,
  pp. 1--7.

\bibitem{ichoua2006exploiting}
S.~Ichoua, M.~Gendreau, J.-Y. Potvin, Exploiting knowledge about future demands
  for real-time vehicle dispatching, Transportation Science 40~(2) (2006)
  211--225.

\bibitem{powell1988comparative}
W.~B. Powell, A comparative review of alternative algorithms for the dynamic
  vehicle allocation problem., in: G.~B. L., A.~A. A. (Eds.), Vehicle Routing:
  Methods and Studies., Elsevier, Amsterdam, The Netherlands, 1988, pp.
  249--291.

\bibitem{powell1996stochastic}
W.~B. Powell, A stochastic formulation of the dynamic assignment problem, with
  an application to truckload motor carriers, Transportation Science 30~(3)
  (1996) 195--219.

\bibitem{godfrey2002adaptive}
G.~A. Godfrey, W.~B. Powell, An adaptive dynamic programming algorithm for
  dynamic fleet management, i: Single period travel times, Transportation
  Science 36~(1) (2002) 21--39.

\bibitem{spivey2004dynamic}
M.~Z. Spivey, W.~B. Powell, The dynamic assignment problem, Transportation
  Science 38~(4) (2004) 399--419.

\bibitem{miller2017predictive}
J.~Miller, J.~P. How, Predictive positioning and quality of service ridesharing
  for campus mobility on demand systems, in: 2017 IEEE International Conference
  on Robotics and Automation (ICRA), IEEE, 2017, pp. 1402--1408.

\bibitem{hall1988distribution}
P.~Hall, S.~J. Sheather, On the distribution of a studentized quantile, Journal
  of the Royal Statistical Society. Series B (Methodological) (1988) 381--391.

\bibitem{snoek2012practical}
J.~Snoek, H.~Larochelle, R.~P. Adams, Practical bayesian optimization of
  machine learning algorithms, in: Advances in neural information processing
  systems, 2012, pp. 2951--2959.

\bibitem{friedman2002stochastic}
J.~H. Friedman, Stochastic gradient boosting, Computational Statistics \& Data
  Analysis 38~(4) (2002) 367--378.

\bibitem{KILIC201421}
F.~Kılıç, M.~Gök,
  \href{http://www.sciencedirect.com/science/article/pii/S0305054814001300}{A
  demand based route generation algorithm for public transit network design},
  Computers \& Operations Research 51 (2014) 21 -- 29.
\newblock \href {http://dx.doi.org/https://doi.org/10.1016/j.cor.2014.05.001}
  {\path{doi:https://doi.org/10.1016/j.cor.2014.05.001}}.
\newline\urlprefix\url{http://www.sciencedirect.com/science/article/pii/S0305054814001300}

\bibitem{hentenryck2009online}
P.~V. Hentenryck, R.~Bent, Online stochastic combinatorial optimization, The
  MIT Press, 2009.

\end{thebibliography}

\end{document}